# Deep Reinforcement Learning for Crowdsourced Urban Delivery: System States Characterization, Heuristics-guided Action Choice, and Rule-Interposing Integration

Tanvir Ahamed[1], Bo Zou[1*], Nahid Parvez Farazi[1], Theja Tulabandhula[2]

[1] Department of Civil, Materials, and Environmental Engineering, University of Illinois at Chicago

[2] Department of Information and Decision Sciences, University of Illinois at Chicago

**Abstract:** This paper investigates the problem of assigning shipping requests to *ad hoc* couriers in the context of crowdsourced urban delivery. The shipping requests are spatially distributed each with a limited time window between the earliest time for pickup and latest time for delivery. The *ad hoc* couriers, termed crowdsourcees, also have limited time availability and carrying capacity. We propose a new deep reinforcement learning (DRL)-based approach to tackling this assignment problem. A deep Q network (DQN) algorithm is trained which entails two salient features of experience replay and target network that enhance the efficiency, convergence, and stability of DRL training. More importantly, this paper makes three methodological contributions: 1) presenting a comprehensive and novel characterization of crowdshipping system states that encompasses spatial-temporal and capacity information of crowdsourcees and requests; 2) embedding heuristics that leverage the information offered by the state representation and are based on intuitive reasoning to guide specific actions to take, to preserve tractability and enhance efficiency of training; and 3) integrating rule-interposing to prevent repeated visiting of the same routes and node sequences during routing improvement, thereby further enhancing the training efficiency by accelerating learning. The effectiveness of the proposed approach is demonstrated through extensive numerical analysis. The results show the benefits brought by the heuristics-guided action choice and rule-interposing in DRL training, and the superiority of the proposed approach over existing heuristics in both solution quality, time, and scalability. Besides the potential to improve the efficiency of crowdshipping operation planning, the proposed approach also provides a new avenue and generic framework for other problems in the vehicle routing context.

---

[*] Corresponding author. Email: bzou@uic.edu.

# 1 Introduction

Urban delivery has been undergoing an exciting and challenging time with ever-growing online shopping demand. In the US, for example, the share of e-commerce sales in total retail sales has increased from 4.2% in the first quarter of 2010 to 16.1% in the second quarter of 2020 (Statista, 2020a). For the food delivery segment alone, revenue amounted to $22.1 billion in 2019 and is projected to reach $32.3 billion by 2024 (Statista, 2020b). This trend has been further accelerated by the COVID-19 pandemic. On the other hand, the time-sensitive nature of today's urban deliveries is imposing considerable pressure on delivery service providers (DSP) to control logistics cost while meeting customer demand for more frequent and expedited deliveries. Guaranteed delivery within one or two hours after an order is placed online has become increasingly common in urban delivery practices (Setzke et al., 2018). As a result of the growing demand and increasing time sensitivity, delivery vehicle traffic has been continuously on the rise in urban areas, causing many negative consequences (Kafle et al., 2017).

To address the challenges, crowdshipping has emerged as an attractive new form of urban delivery. In crowdshipping, a DSP solicits ordinary people, termed crowdsourcees, who have some available time and may walk, bike, or drive a car to perform delivery to earn some payment. Many companies, including Postmates, Deliv, Piggy Baggy, Amazon Flex, Uber Eats, DoorDash, and Instacart, are rapidly expanding their crowdshipping businesses. By using ordinary people as "*ad hoc* couriers", crowdshipping can bring significant cost advantages over the traditional asset-based delivery which relies on full-time employees and self-owned vehicles (Arslan et al., 2019).

In this paper, we focus on crowdshipping with spatially distributed request pickup and delivery locations, using "dedicated crowdsourcees" who are also spatially distributed. Distributed pickup and delivery locations are common for delivery from restaurants, grocery stores, and retail shops to customers, and even for document delivery between different office locations. Dedicated crowdsourcees inform the DSP about their available time for performing delivery, which is a popular form in the crowdshipping business (e.g., Amazon Flex, Instacart, Postmates). This is different from another type of crowdshipping using opportunistic (or "on the way") crowdsourcees: the pickup-delivery tasks are "piggybacked" on the crowdsourcees' existing journey with some extra miles and time incurred. We further consider that each crowdsourcee has a limited carrying capacity. A crowdsourcee is paid a fixed rate ($/minute) whenever the crowdsourcee is carrying a request.

In addition to the description of the type of crowdsourcees, this paper considers a delivery environment where each shipping request has a narrow time window defined by the time between the earliest time of pickup and the latest time of delivery. This characterization is equivalent to specifying that each request needs to be delivered within a certain amount of guaranteed time (e.g., two hours) after the shipping request becomes available. This delivery environment is prevalent today, especially in food and grocery deliveries.

Given the above setup, we consider the following static crowdshipping problem as the central research question of our paper: how can a DSP efficiently assign requests to crowdsourcees by minimizing total shipping cost, while respecting constraints arising from the time availability and carrying capacity limits of crowdsourcees and pickup and delivery time windows of the requests? Cost minimization requires consolidation of multiple requests into one crowdsourcee route, as this can reduce total routing distance thus routing time made by crowdsourcees. In the case that a request is not assigned to a crowdsourcee, the request will have to be picked up and delivered by a backup vehicle, which is more expensive than hiring a crowdsourcee. This crowdshipping problem can be viewed as a specific type of pickup-and-delivery problem and belongs to the broad category of vehicle routing problems (VRP).

While many integer programming models and heuristic algorithms have been developed for solving this and similar problems, the novelty of this paper is that we propose, for the first time in the literature, an approach that leverages deep reinforcement learning (DRL) – more specifically deep Q learning (DQN) – to frame and solve the constrained crowdsourcee-shipping request assignment problem. Two salient features of DQN are experience reply and target network which can enhance efficiency, convergence, and stability in DRL training. Our work goes beyond simple adoption of the DQN algorithm in the existing literature, by making three major methodological contributions as follows.

The first contribution is on a novel representation of system states for the crowdshipping problem. Due to the combinatorial nature of the crowdshipping problem and the heterogeneity of both requests and crowdsourcees in terms of time and carrying capacity, the states of a crowdshipping system cannot be represented by one or a few metrics. A comprehensive representation must in some way capture the sequence of pickup and delivery nodes on each crowdsourcee route. Yet routing sequence alone is not enough to reflect the fact that both requests and crowdsourcees are time sensitive: on the one hand, each request has a limited time window between the earliest possible pickup and the latest delivery (e.g., 2 hours). On the other hand, by dedicating one's time to crowdshipping, a crowdsourcee also has limited time availability. The time information about requests and crowdsourcees, which changes as crowdsourcee routes are constantly created and improved, is an inherent part of the system state that helps the DRL agent make informed routing decisions, especially with respect to what requests need be considered first and what crowdsourcee routes may be given higher priority given time availability and delivery urgency. To this end, a novel representation of system states that leverages the notion of information array is proposed which encompasses not only static location information of request pickup and delivery nodes but information on crowdsourcee routing sequences, request-specific time availability, and crowdsourcee-specific time and capacity availability.

The second contribution is on embedment of heuristics-guided action choice in DRL. The combinatorial nature of the problem means that a very large number of different actions can be taken to construct and improve crowdsourcee routing. But enumerating all possible actions would be neither efficient nor practical in DRL training. To preserve training tractability, we abstract the action space into five general types of actions for assigning or improving the assignment of requests to crowdsourcees: 1) inserting an unassigned request to a crowdsourcee route (insertion); 2) moving an assigned request to another place in the same crowdsourcee route (intra-route move); 3) moving an assigned request to a different crowdsourcee route (inter-route move); 4) exchanging the positions of two requests that are assigned to two different crowdsource routes (1-exchange); 5) do-nothing. As many possibilities for taking a specific action still exist given an action type, heuristics that leverage the information offered by our proposed state representation and are based on intuitive reasonings are designed to guide the specific action to take. We show that the embedment of heuristics-guided action choice significantly enhances DRL training efficiency and solution quality.

The third contribution is on integration of rule-interposing into DRL training and implementation. The rules aim to prevent certain routes or node sequences from being visited repeatedly during neighborhood moves (i.e., intra-route move, inter-route move, and 1-exchange) within a period of time, as repeated visiting discourages exploring more actions and may get the routing sequence trapped in local optimum, thus compromising the efficiency of DRL training. Specifically, we employ two rules that: 1) set up and update a priority list of crowdsourcee routes for each neighborhood move, based on criteria in line with the nature of the neighborhood moves. A crowdsourcee route that is chosen for a neighborhood move will be removed from the priority list and not considered for some period of time; 2) introduce Tabu tenure for the relative positions of pickup and delivery nodes. Two nodes that were neighbored and are moved away are prohibited to be neighbored again for some period of time. With the two rules, computation efforts involved in repeatedly visiting routes or node sequences during neighborhood moves are spared, thereby enhancing the training efficiency by accelerating learning.

With the above three methodological contributions, the effectiveness of the proposed DRL-based approach to solve the crowdshipping problems of our interest is demonstrated through extensive numerical analysis. Our results show superiority of the trained DQN algorithm over traditional heuristics in solution quality, time, and scalability. Given that the training of DRL will be performed offline and a trained DRL model can solve problems in a matter of seconds, the proposed approach has significant potential for practical crowdshipping operation planning and even real-time decision-support. Moreover, the methodology framework, which in this paper tackles a more complicated type of pickup and delivery problems with time constraints from both "vehicles" (crowdsourcees) and "customers" (shipping requests), provides a new and promising avenue and generic framework for solving general pickup and delivery and VRP problems.

The remainder of the paper is structured as follows. Section 2 reviews and synthesizes the relevant literature. In Section 3 provides a detailed presentation of the methodology including the fundamentals of reinforcement learning (RL) and DRL; information array, representation of states, actions, and rewards; the DQN algorithm for crowdshipping; and rule-interposing design. Section 4 implements the DRL model and discusses the results from extensive numerical experiments. Summaries and suggestions for future research are given in Section 5.

## 2 Literature review

We organize our review of the relevant literature in two parts. We first review and synthesize the recent advances of DRL in solving vehicle routing problems including both passenger transportation and freight delivery. We will synthesize the problem characteristics and DRL specifications in representative studies, based on which the uniqueness of our paper is highlighted. We also review the literature of crowdshipping, which reveals that DRL has not been adapted to tackle crowdshipping problems.

### 2.1 DRL for routing problems

#### 2.1.1 Existing studies

The crowdshipping problem studied in this paper is a routing problem, more specifically belonging to the category of pickup and delivery problems. As such problems are NP-hard, previous efforts have been focused on developing efficient heuristics to obtain solutions with acceptable quality and in a reasonable amount of computation time (e.g., Lu and Dessouky, 2006; Berbeglia et al., 2010; Toth and Vigo, 2014). Heuristics are typically expressed in the form of a set of rules, which may be interpreted as policies to make routing decisions (Kool et al., 2018). With the advance in DRL and increasing availability of computation power to researchers in recent years, these policies can be formulated under a reinforcement learning framework, parameterized using deep neural networks (DNN), and trained to obtain new and stronger algorithms. Alone this line of thought, DRL-based solution approaches to solving routing problems are garnering growing attention lately.

A basic version of routing problems is the traveling salesman problem concerning the routing of a single vehicle. Bello et al. (2016) probably make one of the first attempts to combine reinforcement learning with neural networks to tackle traveling salesman problems. A pointer network comprising two recurrent neural networks for encoding and decoding and an attention function is trained with policy gradient. The authors find that the trained DRL model outperforms Christofides heuristics and Google's OR tools for 20, 50, and 100 tasks. Kool et al. (2018) build on Bello et al.'s work by training an attention-based encoder-decoder DRL model. Dai et al. (2017) use a graph embedding network to represent the policy to capture the property of a node in the context of its graph neighborhood. A fitted Q-learning is adopted to learn a greedy policy that is parameterized by the graph embedding network. For TSP problems considered above, only spatial information of nodes and a single vehicle tour are involved. Actions in DRL pertain to adding nodes—one at a time—to progressively construct the vehicle tour.

The complexity of routing problems is augmented when extended to multiple routes, with time constraints, and with pickups and deliveries. Several recent efforts have appeared in the literature, on both freight and passenger sides. For freight delivery problems, Nazari et al. (2018) consider a parameterized stochastic policy to solve a VRP with limited vehicle capacity. The authors apply a policy gradient algorithm to optimize the parameters of the stochastic policy. Chen et al. (2019) use multi-agent RL to train a courier dispatch policy to deal with goods pickups with prescribed pickup time windows. To maintain the state-action space at a controllable size, the RL is decentralized with each courier modeled as an agent. However, the drawback of a decentralized approach is a compromise in modeling the interactions among couriers in undertaking different pickup tasks. The problem considered in Yu et al. (2019) is more similar to what is investigated in our paper. The authors look into a pickup and delivery problem with constraints from vehicle capacity and deadline for each request delivery. Like Chen et al. (2019), the authors opt for a distributed neural optimization strategy where a pointer network with graph embedding is used to progressively develop a complete tour of each vehicle.

On the passenger side, the interest of adopting DRL for VRP arises with the proliferation of ridesharing which involves picking up and dropping off riders. Oda and Joe-Wong (2018) propose a DQN-based framework that learns which zone an idle vehicle should go to. The learning is independent for each vehicle, thus also a distributed dispatching system. A limitation of this study is that each vehicle is assumed to have at most one rider onboard at any point in time. Singh et al. (2019) relax the assumption by allowing more than one rider in a ridesharing vehicle. However, the training remains decentralized, i.e., each vehicle solves its DQN problem without coordination with other ridesharing vehicles in its vicinity. The possibility that a rider may transfer from one vehicle to another vehicle could be an undesirable feature of the proposed service and is actually not commonly seen in ridesharing practice. Another distributed model-free algorithm using DQN techniques that learns optimal dispatch policies for each vehicle individually by distributed training is developed by Al-Abbasi et al. (2019). In that study, the training of a vehicle's dispatching policy again does not consider coordination with other vehicles.

### 2.1.2 Comparison of the existing studies and our work

It can be seen from the above review that most of the multi-vehicle routing problems considered in the existing DRL literature are different in problem characteristics from the crowdshipping problem considered in this paper. As shown in Table 1a, only Al-Abbasi et al. (2019) on the passenger side and Yu et al. (2019) on the freight side consider pickup and delivery with the possibility of a vehicle carrying multiple customers at the same time and without transfer. While vehicle capacity constraints are accounted for in some of the papers, customer time window constraints are mostly not, only in Chen et al. (2019) and Yu et al. (2019). Yet none of the studies considers limited time availability of vehicles, which is an essential characteristic in our crowdshipping problem (where crowdsourcees are "vehicles"). Except for Nazari et al. (2019), in all other works each vehicle is trained individually due to the complexity of DQN training and solutions. In contrast, the full control of a DSP of assigning crowdsourcees to shipping requests means that centralized DQN would be more appropriate. The present paper attempts to overcome these issues while comprehensively capturing the problem characteristics of crowdshipping, as shown in the last row of Table 1a.

Because of the richer features and centralized nature for the crowdshipping system of our interest, fully capturing the states of a crowdshipping system requires more complicated representation than in the existing studies. As shown in Table 1b, the existing studies most have vehicle and/or customer locations as the major part of state representation. Very limited efforts are made to include time-related information for either vehicles or customers. On the other hand, given that both crowdsourcees and shipping requests have limited time windows, and that the DRL proposed in our work embeds heuristics-guided action choice which needs time-related information to proceed, the incorporation of time-related information is critical (see subsections 3.2.2-3.2.3). In fact, the heuristics-guided action choice design preserves training tractability, thereby empowering a centralized policy and contributing to the scalability of the proposed DRL approach. For performing these heuristics, information on routing sequence is needed, which is included in the state representation proposed in this paper, but not in any prior studies reviewed. As a result of the heuristics-embedding feature, the specification of action space in our work is richer and more elaborate. Finally, no other papers consider rule-interposing. These are made clear in Table 1b.

## 2.2 Crowdshipping research

As a result of the rapid development in industry practice, crowdshipping has garnered much research attention in recent years. Relevant literature considers three aspects: supply, demand, and operation and management (Le et al., 2019). As our paper concerns the third aspect, the review in this subsection mainly focuses on different operational schemes and strategies that have been proposed.

Depending on how crowdsourcees are used, existing crowdshipping research can be categorized into two groups. The first group of research considers employing crowdsourcees to perform the last leg of urban delivery, while traditional vehicles are still used for the remaining part. Parcel relay will occur at the interface of the two parts. Wang et al. (2016) formulate a pop station-based crowdshipping problem in which a DSP only sends parcels to a limited number of pop stations in an urban area, while crowdsourcees take care of delivery from the pop stations to customers. Focusing on the crowdsourcee part, a minimum cost network flow model is formulated to minimize crowdsourcing expense while performing the last-leg deliveries. Kafle et al. (2017) investigate relayed crowdshipping system where DSP-owned trucks and crowdsourcees exchange requests at *ad hoc* locations with crowdsourcees taking charge of the last leg of deliveries and the first leg of pickups. A mixed integer program (MIP) along with tailored heuristics is proposed to minimize system overall cost. Macrina et al. (2020) consider that crowdsourcees may pick up parcels from a central depot or an intermediate depot to which parcels are delivered by classic vehicles. Customers can be served either by a classic vehicle or a crowdsourcee. Each crowdsourcee serves at most one customer. To solve this problem, an MIP and a metaheuristic are developed. The use of intermediate transfer locations is further considered by Sampaio et al. (2018) who propose a heuristic for solving multi-depot pickup and delivery problems with time windows and transfers, and by Dötterl et al. (2020) who use an agent-based approach to allow for parcel transfer between crowdsourcees.

The second group of crowdshipping research does not consider intermediate transfers, but uses end-to-end trips either by dedicated crowdsourcees or people "on-the-way" who are willing to make some detour from their original trips to perform delivery. Either way, accounting for the willingness of ordinary people to participate in crowdshipping is important, as considered in Archetti et al. (2016) and Dayarian and Savelsbergh (2017). In these two studies, the matching of crowdsourcees with requests is approached by respectively designing variants of classic capacitated vehicle routing problems and developing two rolling horizon dispatching approaches. A rolling horizon framework with an exact solution approach is proposed by Arslan et al. (2020) which look into dynamic pickups and deliveries using "on-the-way" crowdsourcees. Yildiz and Savelsbergh (2019) introduce the service and capacity planning problem in crowdshipping, and consider the use of both crowdsourced and company-provided capacity to ensure service quality. The possibility of in-store customers to accept and reject a delivery request as a function of the compensation level is incorporated in a bi-level stochastic model in Gdowska et al. (2018). Recognizing that using friends/acquaintances of the shipping request recipients makes delivery more reliable, Devari et al. (2017) conduct a scenario-based analysis to explore the benefits of retail store pickups that rely on friends/acquaintances on their routine trips to stores/work/home. Similarly, Akeb et al. (2018) consider neighbors of shipping request recipients for collecting and delivering requests when the recipients are away from home to reduce failed deliveries.

Despite the proliferation of research in the crowdshipping field, DRL has not been considered in the literature as a way to guide crowdshipping operations. In addition to this major gap, there seems to be little attention paid to the limited time availability that dedicated crowdsourcees are likely to have while performing crowdshipping. This paper intends to address these gaps.

**Table 1a:** Vehicle routing problem characteristics considered in selected DRL studies and the present paper

| | | Problem characteristics | | | | |
|---|---|---|---|---|---|---|
| | | Pickup and delivery | Consider "vehicle" capacity constraint | Considers limited time of "customers" | Considers limited time of "vehicles" | Centralized |
| passenger | Oda and Joe-Wong (2018) | Yes, but one rider in a vehicle at a time | No | No | No | No |
| | Singh et al. (2019) | Yes, but a rider may transfer between vehicles in a trip | Yes | No | No | No |
| | Al-Abbasi et al. (2019) | Yes (ridesharing) | Yes | No | No | No |
| Freight | Nazari et al. (2018) | No | Yes | No | No | Yes (but only 1 vehicle in numerical analysis) |
| | Chen et al. (2019) | No (pickup only) | No | Yes | No | No |
| | Yu et al. (2019) | Yes | Yes | Yes | No | No |
| | **This paper** | **Yes** | **Yes** | **Yes** | **Yes** | **Yes** |

Note: The term "vehicle" is quoted because in crowdshipping, "vehicles" would refer to crowdsourcees. Similarly, the term "customers" is quoted as "customers" would refer to shipping requests on the freight side.

**Table 1b:** DRL specifications in selected studies and the present paper

| | | DRL specification in solving VRP | | |
| --- | --- | --- | --- | --- |
| | | State representation | Action characterization | Rule-interposing |
| Passenger | Oda and Joe-Wong (2018) | 1. Vehicle location<br>2. Occupied/idle status<br>3. Destination of the vehicle<br>4. Number of available vehicles in each zone<br>5. Future demand of each zone | Which zone for an idle vehicle under study to go to | No |
| Passenger | Singh et al. (2019) | 1. Vehicle location (in which zone)<br>2. Available seats of each vehicle<br>3. Rider pickup time<br>4. Rider destination<br>5. Number of vehicles in each zone<br>6. Predicted future rider demand | Which zone to which vehicles are dispatched | No |
| Passenger | Al-Abbasi et al. (2019) | 1. Vehicle location<br>2. Number of available seats<br>3. Rider pickup time<br>4. Rider destination<br>5. When an occupied vehicle becomes available<br>6. Future rider demand | 1. Whether the vehicle under study should pick up new riders<br>2. If yes, which zone to go to | No |
| Freight | Nazari et al. (2018) | 1. Customer location<br>2. Customer demand | Which node to visit by a vehicle | No |
| Freight | Chen et al. (2019) | 1. Number of couriers and requests in each grid<br>2. Total price of requests in each grid<br>3. Distance between neighboring grids<br>4. Score (percent of fulfilled price in total price) | 1. Target grid<br>2. Maximum patrol time in the grid | No |
| Freight | Yu et al. (2019) | 1. Available requests<br>2. Renewable energy generation points<br>3. Next stops of other vehicles in the system<br>4. Battery charging demand of each vehicle | What is the next stop in the tour of the vehicle under study | No |
| | **This paper** | 1. Crowdsourcee starting locations<br>2. Request pickup and delivery locations<br>3. Node precedence relation of crowdsourcee routes<br>4. Request slack time, unused service time, and occupation time<br>5. Crowdsourcee routing duration and remaining available time<br>6. Time and capacity violation of crowdsourcee routes | 1. Inserting a request to a route<br>2. Intra-route move of a request<br>3. Inter-route move of a request<br>4. 1-exchange move of two requests in two routes<br>5. No action | Yes |

# 3 Methodology

This section describes the crowdshipping-adapted DRL methodology. First, we introduce the fundamental ideas of RL and DRL. Then, we discuss how states, actions, and rewards which are essential elements of DRL are specified in crowdshipping. Building on the specifications, we detail the training process using DQN. Two key ideas are worth mentioning. First, DQN learns from how a policy – a decision rule which directs what type of action to take given a state – performed on previous instances and improves the policy over time. Knowing the action type, the specific action will be determined by a corresponding heuristic. As mentioned in subsection 2.1.2, such heuristics-guided action choice preserves training tractability and consequently contributes to the scalability of the proposed approach. Second, solutions to a crowdshipping problem instance can be constructed sequentially, one step at a time, which is amenable to the DRL framework.

## 3.1 Fundamental idea

### 3.1.1 Reinforcement learning (RL)

RL is one of the three categories of machine learning (the other two are supervised learning and unsupervised learning) (Sutton and Barto, 2018). The tenet of RL is to train an agent such that the agent can optimize its behavior by accumulating and learning from its experiences of interacting with the environment. The optimality is measured as maximizing the total reward by taking consecutive actions. Thus, RL is a sequential decision process with the agent as the decision maker. At each decision point, the agent has information about the current state of the environment and selects the best action based on his current experiences. The action taken transitions the environment to a new state. The agent gets some reward, i.e., reinforcement, as a signal of how good or bad the action taken is.

To formulate the sequential decision process, RL employs MDP as the mathematical foundation to keep track of the progression of the decision process. To do so, the following notations are introduced. $S$ is the set of states of the environment. $A$ is the set of actions the agent can take. $R$ is the set of possible rewards as a result of the agent taking an action at a given state. To illustrate, the environment is in state $s_t \in S$ at time step $t$. The agent takes an action $a_t \in A$. The action transitions the environment to a new state $s_{t+1} \in S$ at the next time step $t + 1$. Meanwhile, the agent receives a reward $r_t \in R$. The reward is a function of state-action pair: $r_t(s_t, a_t)$ (Fig. 1).

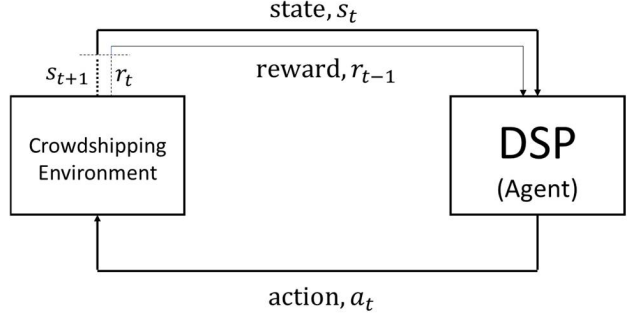

**Fig. 1.** Illustration of states, actions, and rewards

Since the actions are taken sequentially, the objective of the agent at any time step $t$ is to maximize the cumulative reward, i.e., the return $G_t$, from $t$ till the last time step $T$:

$$G_t = r_t + r_{t+1} + \cdots + r_T. \qquad (1)$$

If we consider that the reward is received over a long period, a discount factor $\gamma \in [0,1]$ is often used to reflect discounting:

$$G_t = r_t + \gamma r_{t+1} + \cdots + \gamma^{T-t} r_T \qquad (2)$$

In RL, a policy $\pi$ is a mapping from states to probabilities of selecting each possible action. A value function $V_\pi$ expresses the expected return when starting in state $s$ and following policy $\pi$ thereafter. At time step $t$, the value function can be written as:

$$V_\pi(s) = \mathbb{E}_\pi[G_t | S_t = s] = \mathbb{E}_\pi\left[\sum_{k=0}^{T-t} \gamma^k r_{t+k} \middle| S_t = s\right]. \qquad (3)$$

Related to the value function, we define the value of taking action $a$ in state $s$ and following policy $\pi$ thereafter, denoted as $Q_\pi(s,a)$. $Q_\pi(s,a)$ is termed action-value function, or "Q-function" of the state-action pair $(s,a)$. The letter "Q" represents the quality of this state-action pair:

$$Q_\pi(s,a) = \mathbb{E}_\pi[G_t | S_t = s, a_t = a] = \mathbb{E}_\pi\left[\sum_{k=0}^{T-t} \gamma^k r_{t+k} \middle| S_t = s, a_t = a\right]. \qquad (4)$$

It is desired to seek an optimal policy $\pi^*$ such that $V_{\pi^*}(s) = \max_a Q_*(s, a)$, where $Q_*(s, a)$ means that the agent takes action $a$ at state $s$ and follows policy $\pi^*$ thereafter. Clearly, if $Q_*(s, a)$ is known for every state-action pair $(s, a)$, then $\pi^*$ is also known. The problem of finding the optimal policy then becomes finding optimal Q-values $Q_*(s, a), \forall (s, a) \in S \times A$. To do so, one of the prominent algorithms is Q-learning (Watkins and Dayan, 1992). At a time step, the Q-function value (thereafter simplified as "Q-value") for a given state-action pair is updated using the following rule which is based on the Bellman optimality equation:

$$Q(s, a) \leftarrow (1 - \alpha) Q(s, a) + \alpha \left[ r(s, a) + \gamma \max_{a' \in A} Q(s', a') \right] \quad (5)$$

where $s'$ is the transitioned state after taking action $a$ at state $s$. $r(s, a)$ is the associated reward. On the left-hand side of Eq. (5) is the updated $Q(s, a)$ value. On the right-hand side (RHS), $Q(s, a)$ and $Q(s', a')$ come from the current Q-matrix, which is a mapping from a discrete state-action space to Q-values. $\alpha$ is the learning rate taking values between 0 and 1. It can be shown that Q-learning converges to the optimal Q-values with probability 1 as long as all actions are repeatedly sampled in all states and state-action pairs are discrete (Watkins and Dayan, 1992).

### 3.1.2 Deep reinforcement learning (DRL)

The Q-learning algorithm works well to find the optimal policy when the state-action space is small. However, it would become computationally inefficient and even infeasible to compute Q-values for every state-action pair when the state-action space is large (just imagine Eq. (5) needs to be repeatedly computed for a large number of state-action combinations, with constant updates of the Q-matrix). This is where deep learning can help reduce the computational burden. Specifically, a parameterized DNN can be integrated with an RL algorithm like Q-learning, to efficiently approximate the optimal Q-values instead of maintaining and updating a Q-matrix while applying Eq. (5).

More specifically, we adapt the DQN algorithm, proposed by Minh et al. (2015), to the problem considered in this paper. DQN is a relatively new DRL algorithm that uses a convolutional neural network as a function approximator of the Q-function. DQN has excelled in video game environments where the state-action space is very large. A prominent advantage of DQN is that it overcomes instability and divergence that occur when a nonlinear function approximator such as a neural network is used to represent the Q-function, by embedding two salient feature: experience replay and target network, whose use will be discussed in subsection 3.3. Before getting into the details of DQN, below we first describe our specifications of states, actions, and rewards in the context of crowdshipping.

## 3.2 DRL formulation for crowdshipping

### 3.2.1 Information array

In this section, we propose a novel state and action space design as well as reward function specification for crowdshipping. A key in this proposal is the creation of an information array that contains the routing sequence of each crowdsourcee. Let $J$ and $K$ denote respectively the sets of shipping requests and crowdsourcees. The information array is a $|K| \times (2|J| + 1)$ matrix where $|K|$ and $|J|$ denote respectively the numbers of crowdsourcees (which is equivalent to the number of routes) and shipping requests. Each row indicates the routing sequence of one crowdsourcee. The matrix has $2|J| + 1$ columns to accommodate the extreme possibility that all $|J|$ requests ($2|J|$ nodes) are assigned to a single crowdsourcee plus the origin node of the crowdsourcee (thus one more node needs to be added). For example, if the $k$th row of the information array contains the following tuple: $(u_k, p_1, p_2, d_1, d_2)$, it means that crowdsourcee $k$ will leave his/her origin node $u_k$, go to the pickup node of the first request $p_1$, pick up the second request $p_2$, then drop off the first request $d_1$, and finally drop off the second request $d_2$. In this case, the cells of the first five columns of the $k$th row are occupied, whereas the remaining cells in the row are empty (Fig. 2).

|     | 1       | 2     | 3     | 4     | 5     | ... | $2|J| + 1$ |
|-----|---------|-------|-------|-------|-------|-----|------------|
| 1   | $u_1$   | ...   | ...   | ...   | ...   | ... | ...        |
| ⋮   | ...     | ...   | ...   | ...   | ...   | ... | ...        |
| $k$ | $u_k$   | $p_1$ | $p_2$ | $d_1$ | $d_2$ | 0   | 0          |
| ⋮   | ...     | ...   | ...   | ...   | ...   | ... | ...        |
| $|K|$ | $u_{|K|}$ | ... | ...   | ...   | ...   | ... | ...        |

**Fig. 2.** Illustration of the information array

The information array is constantly updated after every time step. At the beginning, we assume that all requests are assigned to backup vehicles. Thus, initially each row in the information array contains only the origin node of a crowdsourcee.

### 3.2.2 State representation using a three-tuple

The information array provides a foundation for specifying the state space. At each time step $t$, the state of the crowdshipping environment is described by a three-tuple $s_t = \{S^l, S^r, S^c\}$ which provides respectively: 1) location information of pickup and delivery nodes of requests and crowdsourcee routing sequences; 2) request-specific time information; and 3) crowdsourcee-specific time and capacity information. With $\{S^l, S^r, S^c\}$, the agent not only has a complete picture of the crowdsourcee routing sequences, but can leverage the time-related information to perform heuristics-guided actions, as described in subsection 3.2.3.

The first component in the three-tuple, $S^l$, is specified as follows:

$$S^l = \{n_i, n_j^p, n_j^d, n_k^c; \forall i \in J \cup K, j \in J, k \in K\}$$

where

$n_i$  is the coordinate of node $i$;
$n_j^p$  is the coordinate of the successor node of the pickup node of request $j$ if $j$ is assigned;
$n_j^d$  is the coordinate of the predecessor node of the delivery node of request $j$ if $j$ is assigned;
$n_k^c$  is the coordinate of the first node visited by crowdsourcee $k$ if the crowdsourcee is assigned.

Given $n_j^p$, $n_j^d$, and $n_k^c$, the visiting sequence of each crowdsourcee route can be uniquely identified.

The second component in the three-tuple, $S^r$, contains three pieces of request-specific time information:

$$S^r = \{s_j, b_j, o_j; \forall j \in J\}$$

where

$s_j$  is the slack time of request $j$;
$b_j$  is the unused service time of request $j$;
$o_j$  is the occupation time of request $j$.

For a request $j$, slack time $s_j$ measures how urgent it needs to be assigned:

$$s_j = \begin{cases} \left(t_{d_j}^l - t_{p_j}^e\right) - T_{p_j,d_j}^c & f_j = 0 \\ \mathcal{M} & f_j = 1 \end{cases} \quad (6)$$

where

$t_{d_j}^l$  is the latest delivery time for request $j$;
$t_{p_j}^e$  is the earliest pickup time for request $j$;
$T_{p_j,d_j}^c$  is the direct travel time by crowdsourcee from pickup node $p_j$ to delivery node $d_j$;
$\mathcal{M}$  is a very large number;
$f_j$  is a binary variable indicating whether request $j$ is assigned to a crowdsourcee or not.

For an unassigned request $j$, its urgency is the difference between the largest amount of time allowed for pickup and delivery $(t_{d_j}^l - t_{p_j}^e)$, and the minimum amount of time needed to do so by crowdsourcee ($T^c_{p_j,d_j}$). The larger the difference, the lower the urgency with which the request needs to be assigned. For an assigned request, a very large number $\mathcal{M}$ is given, which means that its urgency is effectively zero (as it is already assigned). Using this urgency measure, the agent solving for the assignments can prioritize assigning requests that have not been assigned to crowdsourcees.

The unused service time of a request $j$ ($b_j$) quantifies the gap between the latest delivery time $t^l_{d_j}$ and the actual delivery time $t_{d_j}$ (Eq. (7)). Conceptually, a larger $b_j$ means greater flexibility in altering the way the request is picked up and delivered (e.g., by moving the request to a different position in the assigned crowdsourcee route or to a different route).

$$b_j = t^l_{d_j} - t_{d_j} \tag{7}$$

Note that in the case of an unassigned request, the request will be delivered by a backup vehicle which departs from a pre-specified depot $D$. Assuming that the backup vehicle will leave the depot at the earliest pickup time $t^e_{p_j}$, the actual delivery time will be $t_{d_j} = t^e_{p_j} + T^b_{D,p_j} + T^b_{p_j,d_j}$ where $T^b_{D,p_j}$ and $T^b_{p_j,d_j}$ denote respectively the travel time of the backup vehicle from the depot to the pickup node, and from the pickup node directly to the delivery node.

The occupation time of a request ($o_j$) quantifies the duration between pickup and delivery of a request $j$.

$$o_j = t_{d_j} - t_{p_j} \tag{8}$$

where

$t_{p_j}$     is the pickup time of request $j$ by the assigned crowdsourcee.

For an unassigned request, $t_{p_j}$ is equal to $t^e_{p_j} + T^b_{D,p_j}$. Thus, $o_j = T^b_{p_j,d_j}$.

The third component in the three-tuple, namely $S^c$, contains four pieces of crowdsourcee-specific time and capacity information:

$$S^c = \{d_k, v_k, \tau_k, \eta_k; \forall k \in K\}$$

where

$d_k$     is the routing duration for crowdsourcee $k$;
$v_k$     is the total delivery time violation of requests assigned to crowdsourcee route $k$;
$\tau_k$     is the remaining available time for crowdsourcee $k$;
$\eta_k$     is the total capacity violation along the route of crowdsourcee $k$.

The calculation of $d_k$ is intuitive. $v_k$ is calculated using Eq. (9), where $J^k$ denote the set of requests assigned to crowdsourcee $k$. The max operator is used when delivery is earlier than the latest delivery time (i.e., $t_{d_j} - t_{d_j}^l \leq 0$) such that it does not contribute to the violation:

$$v_k = \sum_{j \in J^k} \max(t_{d_j} - t_{d_j}^l, 0) \tag{9}$$

The remaining available time of crowdsourcee $k$, $\tau_k$, is the difference between the crowdsourcee's total available time ($t_{end}^k - t_{start}^k$) and the route duration ($d_k$), as shown in Eq. (10), where $t_{end}^k$ and $t_{start}^k$ are the end and start of crowdsource $k$'s available time window. An underlying assumption is that an assigned crowdsourcee will start routing at $t_{start}^k$. If the total available time of a crowdsourcee is less than the route duration, $\tau_k < 0$ means that crowdsourcee $k$'s time availability is violated when finishing the last delivery on the route.

$$\tau_k = (t_{end}^k - t_{start}^k) - d_k. \tag{10}$$

Given that a crowdsourcee has limited carrying capacity (measured in weight), the total capacity violation along a crowdsourcee route $\eta_k$ is the total number of capacity violation occurrences at each pickup node:

$$\eta_k = \sum_{j \in J^k} \delta_{p_j} \tag{11}$$

where $\delta_{p_j} = 1$ if the total weight carried right after picking up at node $p_j$ exceeds the carrying capacity, and zero otherwise.

### 3.2.3 Action space design

As mentioned in Section 1, the combinatorial nature of the crowdshipping problem means that a very large number of different actions can be taken to construct and improve crowdsourcee routing. However, enumerating all possible actions would be neither efficient nor practical in DRL training. To preserve training tractability, in this paper we abstract the action space into five types of actions. At each time step, the agent may perform one action from the five types to alter existing crowdsourcee route(s) or create a new crowdsourcee route. The first type of actions pertains to inserting an unassigned request to an existing/new route. The other three types of actions: intra-route move, inter-route move, and 1-exchange, are neighborhood moves of requests that have been previously placed in some existing crowdsourcee routes. The last action type is do-nothing, i.e., no action is taken. We consider do-nothing as an action for preserving good solutions. Specifically, if a very good solution has been achieved, having the option of do-nothing prevents taking another action that would move away from the solution to an inferior solution. Fig. 3 provides an illustration of the first four action types.

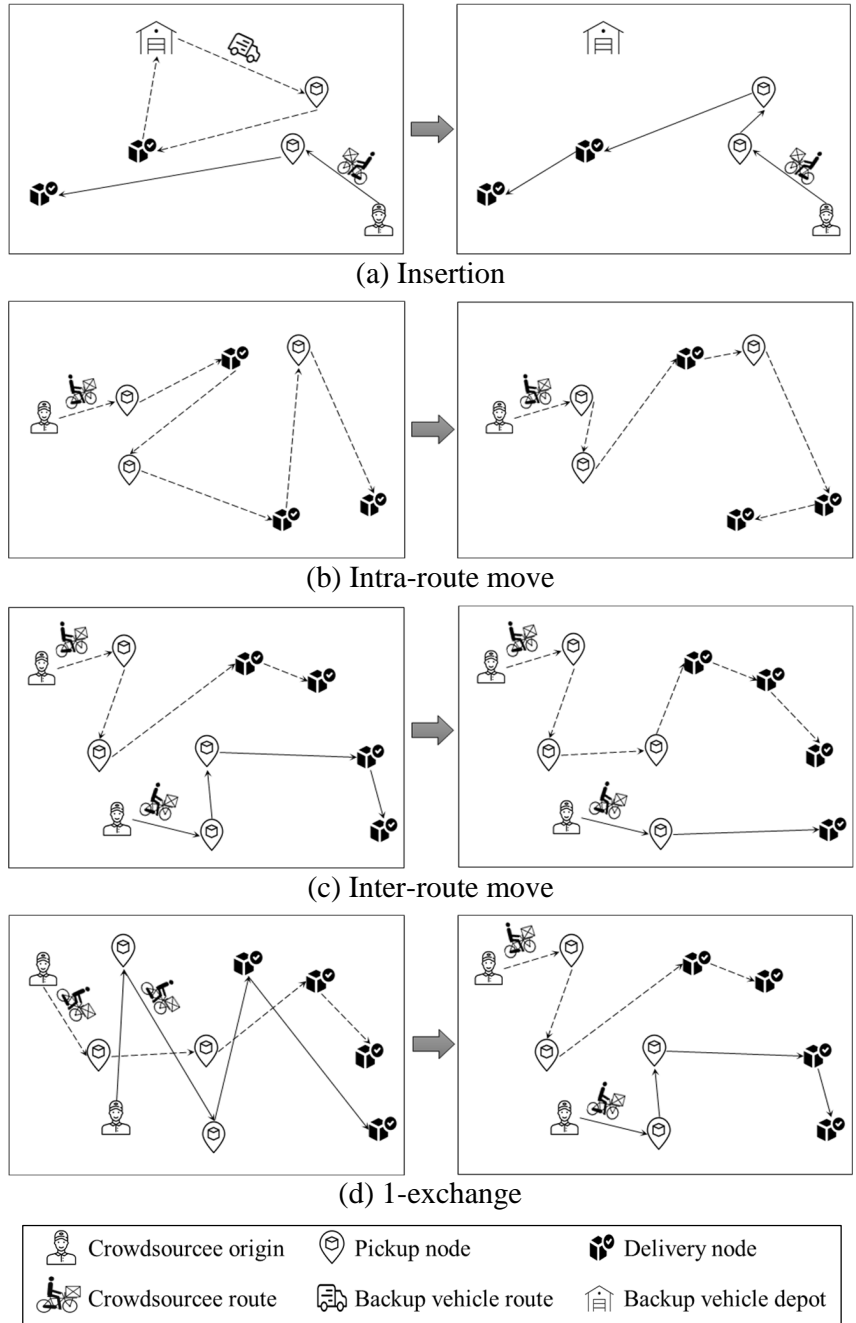

(a) Insertion

(b) Intra-route move

(c) Inter-route move

(d) 1-exchange

**Fig. 3.** Illustration of the four types of actions considered

For insertion, intra-route move, and inter-route move, routing feasibility after taking an action needs to be checked by following Definition 1 below.

**Definition 1.** Feasibility of a crowdsourcee route. A crowdsourcee route $k$ is feasible if the following four conditions are met:

(1) request pickup is no earlier than the earliest pickup, for all requests on the route: $t_{p_j} \geq t_{p_j}^e, \forall j \in J^k$;

(2) request delivery is no later than the latest delivery, for all requests on the route: $t_{d_j} \leq t_{d_j}^l, \forall j \in J^k$;

(3) remaining available time of the crowdsourcee after completing the route is non-negative: $\tau_k \geq 0$;

(4) no violation of crowdsourcee capacity on the route: $\eta_k = 0$.

The remainder of this subsection describes, after an action type is chosen (by the DQN algorithm), how the specific action to take will be identified based on system state information and intuitive reasonings.

*3.2.3.1 Inserting an unassigned request in an existing/new route*

For insertion, we need to determine which request to choose for insertion, and where to insert the request. The action consists of three steps.

---

**Step 1:** *Select a request.*

Among the unassigned requests, select one with the smallest slack time.

**Step 2:** *Insert the request to a route.*

For the selected request, calculate the distances between the pickup node of the request and each crowdsourcee. For an assigned crowdsourcee, the distance is to the end of the crowdsourcee route. For an unassigned crowdsourcee, the distance is to the crowdsourcee origin. Identify the smallest distance.

If the smallest distance occurs to an assigned crowdsourcee, insert the node to the end of the crowdsourcee route. If the smallest distance occurs to an idle crowdsourcee, create a new route: crowdsourcee origin → request pickup node → request delivery node.

Check routing feasibility. If not feasible, move to the crowdsourcee with the second smallest distance and check feasibility. If it is not possible to feasibly insert a request, then move to the request with the second smallest slack time, until a successful insertion.

**Step 3**: *Perform intra-route move.*

If the request is inserted to an existing crowdsourcee route, explore moving the pickup and delivery nodes of the request to earlier positions in the route to reduce routing cost. The move follows Step 2 of intra-route move below. Place the request at the position that leads to the smallest routing cost.

---

In Step 1, the rationale for considering the unassigned request with the smallest slack time, the information of which comes from $s_j$ in $S^r$ (second component in the three-tuple state representation), is that we want to get the most urgent unassigned request assigned first. In Step 2, we perform insertion to the nearest crowdsourcee as this incurs the smallest time loss between the crowdsourcee finishing the currently assigned requests and picking up the request under study.

*3.2.3.2  Intra-route move*

Intra-route move involves moving a later request to an earlier position in a route to reduce routing cost. The action also consists of three steps.

> **Step 1:** *Select a route.*
>
> Select the crowdsourcee route with the largest remaining available time.
>
> **Step 2:** *Move examination.*
>
> Enumerate all feasible moves of a request to an earlier place. To illustrate, consider routing sequence $(u_k, p_1, d_1, p_2, d_2, \ldots, p_{n-1}, d_{n-1}, p_n, d_n)$ and moving request $n$. Move $p_n$ to an earlier position, one place at a time, i.e., to the places right before $d_{n-1}$, right before $p_{n-1}$,…, until right after $u_k$. For each new position of $p_n$, examine feasibility of holding $d_n$ at its initial place, moving it one place at a time to an earlier position, as long as $d_n$ is not before $p_n$. For each feasible $(p_n, d_n)$ move, calculate the routing cost.
>
> Repeat the above move for every request in the route.
>
> **Step 3:** *Identify the best move.*
>
> Among all the feasible moves in Step 2, pick the one with the smallest routing cost. If the routing cost is smaller than the original routing cost, perform the move.

In Step 1, the rationale for considering the route with the largest remaining available time, for which the information comes from $\tau_k$ in $S^c$, is that such a route has the greatest flexibility for moving requests around.

*3.2.3.3  Inter-route move*

Inter-route move picks a request from a route and moves it to another route by performing the following three steps.

| **Step 1:** | *Select a request.* |
| --- | --- |
| | Among all assigned requests, select one with the largest occupation time. |
| **Step 2:** | *Move the request to the end of a different route.* |
| | Insert request to another existing route or create a new route, following Step 2 of insertion. |
| **Step 3:** | *Perform intra-route move of the request.* |
| | Follow Step 3 of insertion. |

In Step 1, the rationale for considering the request with the largest occupation time, for which the information comes from $o_j$ in $S^r$, is that larger occupation time may suggest greater time (and thus cost) reduction potential by moving the request to a different route.

### 3.2.3.4  1-exchange move

1-exchange move pertains to exchanging two requests which are on two crowdsourcee routes. Performing the move has four steps.

| **Step 1:** | *Select the first request.* |
| --- | --- |
| | Among all assigned requests, select the first request that has the largest unused service time. |
| **Step 2:** | *Select the second request.* |
| | Excluding the route associated with the first selected request, select the second request that has the largest unused service time among the remaining assigned requests. |
| **Step 3:** | *Exchange the selected requests.* |
| | Remove the two requests from their routes. Add each request to the end of the other route. |
| **Step 4:** | *Perform intra-route move of the two requests.* |
| | Perform intra-route move for the two requests following Step 3 of insertion. |

In Steps 1 and 2, the rationale for choosing requests with the largest unused service time, for which the information comes from $b_j$ in $S^r$, is that such requests have the greatest flexibility to be moved around. It should be noted that unlike the other three actions, we do not require feasibility to be met in performing 1-exchange (so strictly speaking, the intra-route move in Step 4 here is a modified version of Step 3 of insertion, without checking feasibility). This is intentional to help the search escape local optima (Nanry and Barnes, 2000).

### 3.2.4 Reward specification

Given the state and the action at a time step, we specify the reward as the change in Total Shipping Cost (TSC) as a result of the action taken. If the action taken at time step $t$ is inserting request $j$ in crowdsourcee route $k$, the reward is computed as:

$$r_t = \beta^c d_{k,t-1} + \beta^b \left(T^b_{D,p_j} + T^b_{p_j,d_j} + T^b_{d_j,D}\right) - \beta^c d_{k,t} \tag{12}$$

where

| | |
|---|---|
| $\beta^b$ | is the unit cost of using a backup vehicle (in \$/minute); |
| $\beta^c$ | is the unit cost of using a crowdsourcee (in \$/minute); |
| $d_{k,t-1}$ | is the route duration (in minutes) of crowdsourcee route $k$ at time step $t-1$; |
| $d_{k,t}$ | is the route duration (in minutes) of crowdsourcee route $k$ at time step $t$; |
| $T^b_{d_j,D}$ | is the backup vehicle travel time from the delivery node of request $j$ back to depot $D$. |

In Eq. (12), $\beta^c d_{k,t-1}$ is the cost of crowdsourcee route $k$ at time step $t-1$, which is before request $j$ is inserted. If the route does not exist before inserting $j$, this term will be zero. $\beta^b (T^b_{D,p_j} + T^b_{p_j,d_j} + T^b_{d_j,D})$ is the cost of picking up and delivering the request by a backup vehicle. $\beta^c d_{k,t}$ is the cost of the crowdsourcee route $k$ at time step $t$, after request $j$ is inserted.

If the action taken at time step $t$ is a neighborhood move, let us use $\Psi_t$ to denote the set of crowdsourcee route(s) that are involved in the move. For intra-route move, $\Psi_t$ will have just one route. For inter-route move and 1-exchange, $\Psi_t$ will have two routes. The reward is calculated as the difference of the routing costs before and after the move:

$$r_t = c_t^1 - c_t^2 \tag{13}$$

where $c_t^1$ and $c_t^2$ are routing costs for the route(s) in $\Psi_t$ before and after the neighborhood move:

$$c_t^1 = \beta^c \left( \sum_{k \in \Psi_t} d_{k,t-1} + \vartheta \sum_{k \in \Psi_t} v_{k,t-1} + \tau \sum_{k \in \Psi_t} \chi_{k,t-1} + \rho\phi \sum_{k \in \Psi_t} \eta_{k,t-1} \right) \tag{14}$$

$$c_t^2 = \beta^c \left( \sum_{k \in \Psi_t} d_{k,t} + \vartheta \sum_{k \in \Psi_t} v_{k,t} + \tau \sum_{k \in \Psi_t} \chi_{k,t} + \rho\phi \sum_{k \in \Psi_t} \eta_{k,t} \right) \tag{15}$$

where

| | |
|---|---|
| $d_{k,t}$ | is the route duration for crowdsourcee $k$ at time step $t$; |
| $v_{k,t}$ | is the delivery time violation of requests assigned to crowdsourcee route $k$ at time step $t$; |
| $\chi_{k,t}$ | is the violation of available time for crowdsourcee $k$ at time step $t$: $\chi_{k,t} = \min(\eta_{k,t}, 0)$; |
| $\eta_{k,t}$ | is the carrying capacity violation for crowdsourcee $k$ at time step $t$; |
| $\vartheta$ | is the penalty for delivery time violation; |
| $\tau$ | is the penalty for crowdsourcee overworking; |
| $\rho$ | is the penalty for crowdsourcee carrying capacity violation; |
| $\phi$ | is the capacity violation-to-time conversion factor. |

## 3.3 DRL algorithm for crowdshipping

This subsection describes how the DQN algorithm is adopted to the crowdshipping context. In DQN, the training of the agent is through multiple episodes. Each episode is associated with a different crowdshipping problem instance of a certain size, which is randomly generated and starts with an initial state that all crowdsourcees are idle (unassigned). Training in an episode involves improving the solution to the problem by taking actions described in subsection 3.2.3, one at a time in a number of time steps.

At each time step, an $\varepsilon$-greedy strategy is employed to consider both exploration and exploitation as the agent decides what type of action to take among insertion, intra-route move, inter-route move, 1-exchange move, and do-nothing. By exploration, it means that the agent takes a random action type, with probability $\varepsilon$. By exploitation, the agent takes one of the five action types above that is the best – based on the experiences that the agent has learned so far (reflected in the current Q-values, as shown in line 7 in Algorithm 1 at the end of this subsection), with probability $1 - \varepsilon$. Once the best action type is chosen, the specific action follows the heuristics described in subsection 3.2.3 (line 8 in Algorithm 1). Consequently, a reward and a new state are observed.

While exploitation takes advantage of what have been learned in terms of the best action to take, exploration is necessary to try to get the agent out of local optima toward even better action sequences, to further reduce total shipping cost. At the beginning of an episode, $\varepsilon$ takes value 1, i.e., the focus is purely on exploration, which is intuitive as the agent has zero learned experience (thus nothing to exploit) at this point. Then as time goes by, the agent gradually increases the probability of exploiting learned actions. A decay rate of $\xi$ is used which describes the change in probabilities between two time steps (Eq. (16)). $\xi$ is a hyper parameter.

$$\varepsilon_{t+1} = \varepsilon_t(1 - \xi) \tag{16}$$

One salient feature of DQN is experience replay, for which a replay memory $M$ is used to store the agent's experiences during training. Up to $|M|$ experiences can be stored in the replay memory. An experience is associated with taking an action at a given state and time step, observing a state transition, and getting a reward. For example, at time step $t$, the agent performs an action $a_t$ which transforms the state from $s_t$ to $s_{t+1}$ and yields a reward $r_t$. The experience is denoted as $e_t = (s_t, a_t, r_t, s_{t+1})$. At the beginning of the training, $M$ is empty. As the training continues, experiences are accumulated and added to replay memory $M$. Once $|M|$ experiences are stored in $M$, adding a new experience requires simultaneous removal of the oldest experience stored in $M$.

At each time step, a DNN is trained using a minibatch $M_{\text{sub}}$ of samples that are randomly selected from $M$. Note that in the beginning of the training, the number of accumulated experiences in $M$ will be fewer than $|M_{\text{sub}}|$. In this case, experiences will continuously be accumulated in $M$ but DNN will not be trained, until the replay memory has $|M_{\text{sub}}|$ experiences. The employment of experience replay using randomly selected minibatch samples has multiple advantages. First, because the samples are randomly selected, correlation between samples will be less compared to learning directly from consecutive samples, thereby enhancing the efficiency of learning. Second, each experience can potentially be used in many weight updates, thus allowing for greater data efficiency. Third, by experience replay the behavior distribution is averaged over many previous states, which contributes to smoothing out learning and avoiding oscillation or divergence in the parameters (Mnih et al., 2015).

For each selected experience $(s, a, r, s')$, state $s$ is used as the input for the DNN (with weight parameters $\theta$) to generate state-action value $Q(s, a: \theta)$, or Q-value, which is the output of the DNN.[2] Collectively for all the selected experiences, the prediction of $Q(s, a: \theta)$'s comprises the first *forward pass*. $Q(s, a: \theta)$ is then compared with the target optimal Q-value $Q_*(s, a)$ which gives the maximum expected return achievable by following any DQN policy. Ideally, the target optimal Q-value should satisfy the Bellman optimality equation:

$$Q_*(s, a) = \mathbb{E}_{s'}\left[r + \gamma \max_{a' \in A} Q_*(s', a') \Big| s, a\right] \qquad (17)$$

where $r$ is the immediate reward by taking action $a$ at station $s$.

---

[2] In this paper, we use an architecture in which there is a separate output for each possible action. Only the state is the input to the DNN. Thus, among the outputs for different actions, we choose the one corresponding to action $a$ as $Q(s, a: \theta)$.

The comparison of $Q(s,a:\theta)$ with $Q_*(s,a)$ is performed using a loss function. Assuming a square form for the loss function and replacing $Q_*(s,a)$ by the RHS of Eq. (17), the loss function $\mathcal{L}$, which depends on DNN weight parameters $\theta$, can be expressed as:

$$\mathcal{L}(\theta) = \frac{1}{|M_{\text{sub}}|} \sum_{(s,a,r,s') \in M_{\text{sub}}} \left[ r + \gamma \max_{a' \in A} Q_*(s',a') - Q(s,a:\theta) \right]^2 \quad (18)$$

Obviously, to calculate $\mathcal{L}(\theta)$, $Q_*(s',a')$ is needed. However, $Q_*(s',a')$ is unknown (if $Q_*(s',a')$ was known, then the training would be done). One way to get an approximation of $Q_*(s',a')$ is to perform another *forward pass* with the DNN, i.e., for state $s'$ in each experience $(s,a,r,s')$ along with the same weight parameters $\theta$ of the DNN, predict state-action values $Q(s',a':\theta), \forall a' \in A$ using the DNN. By approximating $Q_*(s',a')$ with $Q(s',a':\theta)$, Eq. (17) can be re-expressed as:

$$\mathcal{L}(\theta) = \frac{1}{|M_{\text{sub}}|} \sum_{(s,a,r,s') \in M_{\text{sub}}} \left[ r + \gamma \max_{a' \in A} Q(s',a':\theta) - Q(s,a:\theta) \right]^2 \quad (19)$$

After performing two forward passes as described above, the gradient of the loss in Eq. (19) is used to update $\theta$ by the Adam optimizer, a widely-used gradient descent-based algorithm for minimizing the loss (Kingma and Ba, 2015). However, a main drawback exists in this two-forward pass procedure. When $\theta$ gets updated, the Q-values obtained from this network will also get updated (in the next time step). So will the target Q-values as they are calculated using the same network parameter. In other words, the direction of updates for the Q-values and the target Q-values will be same. As a consequence, the correlation between the Q-values and the target Q-values can be high, possibly leading to oscillation or divergence of the policy during training.

To tackle this issue, a second salient feature of DQN is that a parallel network, called the target network, of the original DNN is created to preserve DNN parameter values for a period of time, so that target Q-values do not get updated with the same frequency as the Q-values. The target network, which is a clone of the original network, initializes its parameters $\theta'$ using the original DNN: $\theta' = \theta$ at the beginning of the training. Then, instead of updating $\theta'$ by $\theta$ of the original DNN at every time step, $\theta'$ is frozen for $\delta$ time steps. Only after every $\delta$ time steps, $\theta'$ gets updated to whatever is the present value of the original network parameters $\theta$. In this procedure, $\delta$ is a hyper parameter.

The Q-value obtained from the target network $Q(s',a':\theta')$ is used to calculate the approximate target Q-value $r + \gamma \max_{a' \in A} Q(s',a':\theta')$. The loss function shown in Eq. (17) becomes:

$$\mathcal{L}(\theta) = \frac{1}{|M_{\text{sub}}|} \sum_{(s,a,r,s') \in M_{\text{sub}}} \left[ r + \gamma \max_{a' \in A} Q(s', a': \theta') - Q(s, a: \theta) \right]^2 \tag{20}$$

In implementing DQN, we use a slightly modified version of the squared loss function called Huber loss function. For each sample, the squared term is used only if the absolute error falls below a threshold (here we choose the value 1). Otherwise, we use an absolute term as shown in Eq. (21). An advantage of the Huber function form is that the loss is less sensitive to outliers than the square loss for large errors, which prevents exploding gradients.

$$\mathcal{L}(\theta) = \frac{1}{|M_{\text{sub}}|} \sum_{(s,a,r,s') \in M_{\text{sub}}} L_H \left( r + \gamma \max_{a' \in A} Q(s', a': \theta') - Q(s, a: \theta) \right) \tag{21}$$

where

$$L_H \left( r + \gamma \max_{a' \in A} Q(s', a': \theta') - Q(s, a: \theta) \right) = \begin{cases} 0.5 \left[ r + \gamma \max_{a' \in A} Q(s', a': \theta') - Q(s, a: \theta) \right]^2 \\ \quad \text{if } \left| r + \gamma \max_{a' \in A} Q(s', a': \theta') - Q(s, a: \theta) \right| < 1 \\ \left| r + \gamma \max_{a' \in A} Q(s', a': \theta') - Q(s, a: \theta) \right| - 0.5 \text{ otherwise} \end{cases}$$

Finally, it should be mentioned that during an episode, we also accumulate the rewards that are negative. If the accumulated negative reward in an episode falls below a threshold, then the training of the episode is perceived as not promising and consequently terminates.

Summarizing, the overall learning algorithm is presented in Algorithm 1 and illustrated in Fig. 4 below.

**Algorithm 1:** DQN algorithm used in the crowdshipping problem
1. Initialize replay memory $M = \emptyset$
2. Initialize the original DNN with random weight parameters $\theta$
3. Initialize the target DNN with same structure as the original DNN and weight parameters $\theta' = \theta$
4. **for** episode $i = 1$ to $I$, **do**  ▷ $I$ is the number of episodes
5.     Initialize state $s_0 \in S$  ▷ in the initial state $s_0$, all crowdsourcees are unassigned
6.     **for** time step $t = 1$ to $T$, **do**  ▷ $T$ is the number of time steps in an episode
7.         Select a random action type $a_t$ with probability $\varepsilon$; otherwise, set action type $a_t = \underset{a \in A}{\mathrm{argmax}}\, Q(s_t, a; \theta)$
8.         Execute a specific action under action type $a_t$, as guided by the heuristics in subsection 3.2.3. This results in $r_t$ and $s_{t+1}$
9.         Store experience $e_t = (s_t, a_t, r_t, s_{t+1})$ in $M$
10.        **if** $|M| > |M_{\mathrm{sub}}|$, **do**
11.            **if** $\mathbb{R}_t > \mathcal{K}$, **do**  ▷ $\mathbb{R}_t$ is accumulated negative reward in the episode; $\mathcal{K}$ is a threshold
12.               Randomly sample a minibatch $M_{\mathrm{sub}}$ of experiences from $M$
13.               **for** each experience $e_j = (s_j, a_j, r_j, s_{j+1})$ in $M_{\mathrm{sub}}$, **do**
14.                   Compute $r_j + \gamma \underset{a' \in A}{\max}\, Q(s_{j+1}, a' : \theta')$
15.               **end for**
16.               Calculate loss by Eq. (21)
17.               Update weight parameters $\theta$ by the Adam optimizer
18.               Update $\theta' = \theta$ every $\delta$ time steps
19.            **else**
20.               **break**  ▷ if $\mathbb{R}_t < \mathcal{K}$, the training is perceived as not promising and stop
21.            **end if**
22.         **end if**
23.     **end for**
24. **end for**

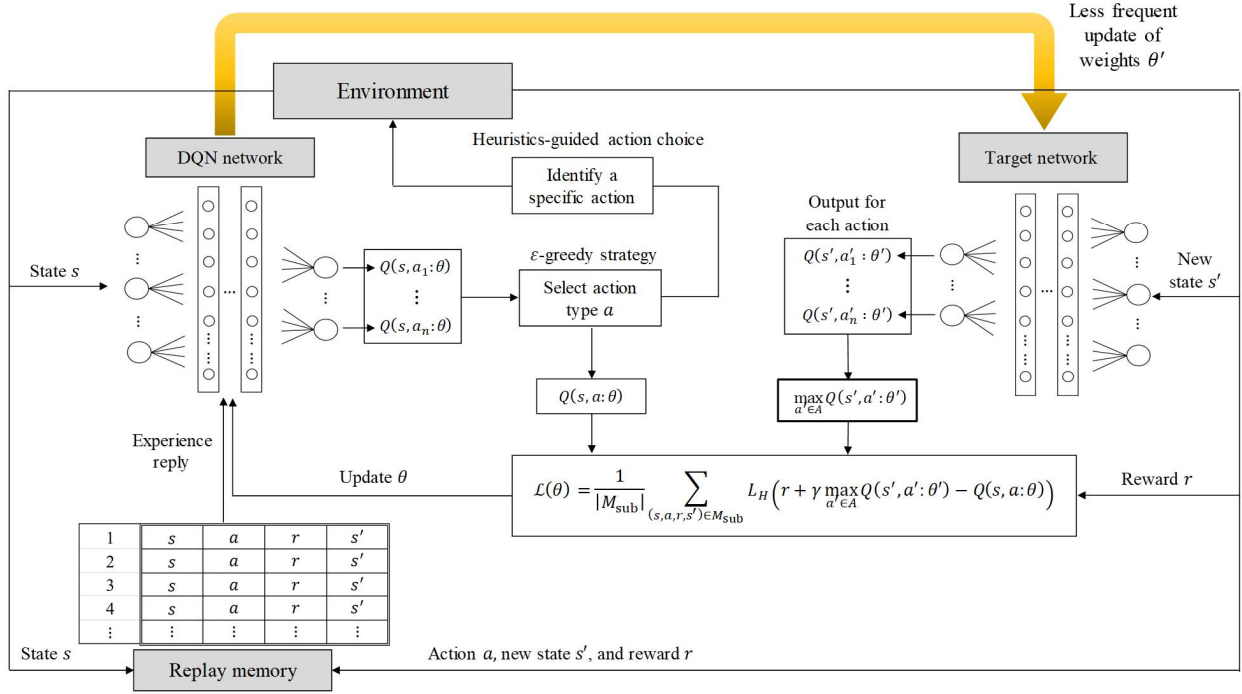

**Fig. 4.** The architecture of the DQN algorithm

### 3.4 Rule-interposing in DRL training and implementation

Whether in DRL training or in implementation of the trained policy, it is possible that some routes or node sequences are repeatedly visited during neighborhood moves. This reduces the efficiency of DRL training, as well as the efficiency in search for the best crowdsourcee-request assignment outcome when a trained policy is applied to solve a problem instance (note that after DRL training is done, at a given state $s$ the optimal Q-value only provides what type of action to take, i.e., $a^* = \underset{a \in A}{\mathrm{argmax}}\, Q^*(s, a)$ where $Q^*$ denotes the optimal Q-value). In this subsection, we propose two rules that aim to prevent such repeated visiting of routes and node sequences, by excluding a previously visited route or node sequence from being considered again in a number of subsequent actions. In what follows, the first rule focuses on routes. The second rule focuses on node sequences.

#### 3.4.1 Rule 1: Introducing priority lists for route selection

To avoid that actions are repeatedly exerted on one or a subset of crowdsourcee routes, the first rule proposed relies on construction and use of three priority lists of crowdsourcee routes, with each list corresponding to one of the three neighborhood move action types (intra-route move, inter-route move, and 1-exchange) described in subsection 3.2.3. Specifically, when a neighborhood move action type is chosen, we pick the crowdsourcee route(s) from the top of the corresponding priority list to apply the action type. After the specific action is taken on the route(s), the route(s) are removed from the list. Thus over time, routes will be continuously picked and removed from the priority list. The priority list will be shortened, and eventually become empty. Then, we construct a new priority list of all the crowdsourcee routes for the same action type. By doing so, during the life cycle of a priority list, a route is considered only once for the associated action type. This allows more exploration of the same action type on other routes. This construction-destruction of priority lists repeats throughout the training and implementation of a trained DRL to solve a problem instance.

Each of the three priority lists is constructed based on some criterion. For intra-route move, the priority list is constructed by sorting crowdsourcee routes in descending order based on the crowdsourcee's remaining available time, which is consistent with the rationale of Step 1 in subsection 3.2.3.2. For inter-route move, the priority list is constructed by sorting crowdsourcee routes in descending order based on the occupation time of each crowdsourcee, measured as the duration between the time of the last delivery and the time of the first pickup. This is in line with Step 1 in subsection 3.2.3.3 (there we also consider the largest occupation time, though for requests). Thus, request selection of Step 1 in subsection 3.2.3.3 will be only from the route with the highest priority. After an inter-route move action is taken, that route is removed from the priority list. The occupation time of the route to which the request is moved will be updated. The position of that route in the priority list will also be updated, for which the computational complexity is $O(\log(N))$ based on binary search, with $N$ being the number of crowdsourcee routes in the priority list. For 1-exchange move, the priority list is constructed by sorting crowdsourcee routes in descending order based on unused service time, which is consistent with Steps 1 and 2 in subsection 3.2.3.4. Thus, the selection of the first and the second requests will be from the two routes with the highest and second highest priority respectively. After a 1-exchange move action is taken, the two routes will be removed from the priority list.

### 3.4.2 Rule 2: Imposing Tabu tenure for neighborhood moves

The second interposing rule is that after a request node (either pickup or delivery) is moved away from a neighboring node (either right before or right after in the routing sequence) on a crowdsourcee route, the former node cannot be moved back to the same location relative to the latter node over a certain number of subsequent actions. This latter node can be of a different request, or of the same request (i.e., the former node is the pickup node of a request, and the latter node is the delivery node of the same request). Similar to Rule 1, this rule applies to the three types of neighborhood moves (intra-route move, inter-route move, and 1-exchange). For each type of neighborhood move, a Tabu tenure will be created to record for how many subsequent actions a request node cannot be neighbored with another node. Similar to Rule 1, Tabu tenure allows neighborhood moves to explore more routing sequences, rather than getting trapped in routing sequences that have been explored and only locally optimal.

To operationalize Tabu tenure, two matrices are created and maintained. The first matrix, of dimension $2|J| \times 2|J|$, indicates whether a node (indexed by the column. There are in total $|J|$ requests thus $2|J|$ pickup and delivery nodes) preceding another node (indexed by the row) is Tabu-ed and for how long. The second matrix, of dimension $(|K| + 2|J|) \times 2|J|$, indicates whether a node (indexed by the column) following another node (indexed by the row) is Tabu-ed and for how long. The second matrix has $|K|$ more rows which correspond to the origins of the $|K|$ crowdsourcees, as a pickup node can be placed right after the origin of a crowdsourcee. These two matrices are updated whenever an action is taken. Note that if a Tabu-ed position yields a solution that is better than the best solution obtained so far for a problem, then the Tabu tenure will be overridden.

## 4 Numerical experiments

This section illustrates numerical implementation of the DQN algorithm described in Section 3, in two problem sizes: one of a medium size with 50 requests and 22 crowdsourcees, and the other of a larger size with 200 requests and 70 crowdsourcees. We present and discuss the results for medium-size problems in great detail, including problem setup, training results, assessment of the benefits of heuristics-guided action choice and rule-interposing, comparison with three popular heuristic methods, and results sensitivity to key hyperparameters. For the larger-size problems, for paper length considerations we report more briefly the results of implanting the trained DRL model to 20 different problem instances in terms of total shipping cost and computation time, in comparison with the three heuristics. The DQN algorithm is coded and trained in the PyTorch environment. All numerical investigations are conducted on a personal computer with Intel Core i9-10920X CPUs at 3.50GHz and 128GB RAM and NVIDIA Titan RTX GPUs.

### 4.1 Medium-size problems

#### 4.1.1 Setup

As mentioned above, we consider a static problem of assigning 50 requests to 22 crowdsourcees. The service area has a square shape of 6 miles × 6 miles. The pickup and delivery locations of each request is randomly generated in the service area. So are the origins of the crowdsourcees. We assume that crowdsourcees bike to perform pickup and delivery at a speed of 10 mph. The carrying capacity of a crowdsourcee is 10 lbs. The available time of a crowdsourcee is randomly drawn from a uniform distribution of 1-2 hours. The weight of a shipping request is also randomly drawn from a uniform distribution of 2-7 lbs. The early pickup time of all requests is the present time. The latest delivery time of a request is randomly drawn from a uniform distribution of 100-120 minutes. If a request is not assigned to any crowdsourcees, it will be picked up and delivered by a backup vehicle which leaves a depot located at the center of the service area and returns to the depot after finishing the delivery. Given the small weight of a request relative to the typical carrying capacity of a backup vehicle, capacity constraints are not considered for backup vehicles. We assume backup vehicles travel at a speed of 20 mph. We follow Kafle et al. (2017) by setting the operating cost of a backup vehicle to be \$68/hour (\$1.13/minute) and the pay rate for crowdsourcees to be \$10/hour (\$0.17/minute), which is considerably cheaper. Crowdsourcees get paid whenever carrying requests.

Following Mnih et al. (2015), values of hyperparameters, shown in Table 1, are selected by performing an informal search. We set the length of an episode to be 85 time steps, the penalty parameters in the reward specification to be $\vartheta = 0.1$, $\tau = 0.2$, and $\rho\phi = 0.15$, and the length of Tabu tenure to be three subsequent actions. We choose a 5-layer fully connected feed-forward neural network as our DNN construction, where each hidden layer has 128 neurons.

Table 1: Hyperparameters values

| Hyperparameter | Value |
|---|---|
| Replay memory size ($|M|$) | 10,000 |
| Minibatch size ($|M_{\text{sub}}|$) | 100 |
| Target network update frequency ($\delta$) | 400 |
| Discount factor ($\gamma$) | 0.96 |
| Learning rate ($\alpha$) | 0.001 |
| Decay rate ($\xi$) | 0.001 |
| Episode termination threshold ($\mathcal{K}$) | -25 |

### 4.1.2 Training results

Fig. 5 plots the evolution of training over time steps, using four measures: (a) average loss per episode. Thus each point is associated with an episode; (b) average Q-value, averaged over all state-action pairs in a minibatch. In this figure, the light purple curves reflect the actual values, while the red curve is the running average, averaged over 65 time steps; (c) accumulated reward between two terminations. As described in subsection 3.3, a termination occurs when the accumulated negative reward falls below threshold $\mathcal{K}$ (line 20 of Algorithm 1); (d) cumulative penalty, which is the sum of the last three terms in the parentheses (multiplied by $\beta^c$) in Eq. (14)-(15) over all time steps from the start of the training. We keep training the DQN algorithm and track the relative change for all the curves (a)-(d) in the most recent 3,000 time steps. The training stops when the relative change for cumulative penalty is less than 5%. In total, 47,039 time steps are used in the DQN training.

Fig. 5(a) shows that the average loss per episode follows an increasing trend up to around 20,000 time steps and tends to stabilize afterwards. In Fig. 5(b), the average Q-value keeps improving till after 30,000 time steps. Before that, updates in the DNN considerably improve the DQN algorithm which yields better solutions than before each update. As a result, the average Q-value keeps improving. Fig. 5(b) also shows a magnifier of the first 5,000 time steps. It is interesting to observe step-wise jumps almost every 400 time steps, which is the target network update frequency (see Table 4). In other words, whenever an update of the target network occurs, it improves the average Q-value. The magnitude of the jumps decreases over time steps, suggesting that the marginal improvement of the DQN algorithm from training is diminishing as training continues. In Fig. 5(c), the demarcation between unstable and stable accumulated reward between two terminations is also around 30,000 time steps. Fig. 5(d) shows that the accumulative penalty over all time steps becomes stable a bit later: after around 40,000 time steps, the DQN algorithm becomes well trained that taking actions suggested by the DQN algorithm will cause little violation of time and capacity constraints (which incurs penalty) during neighborhood moves.

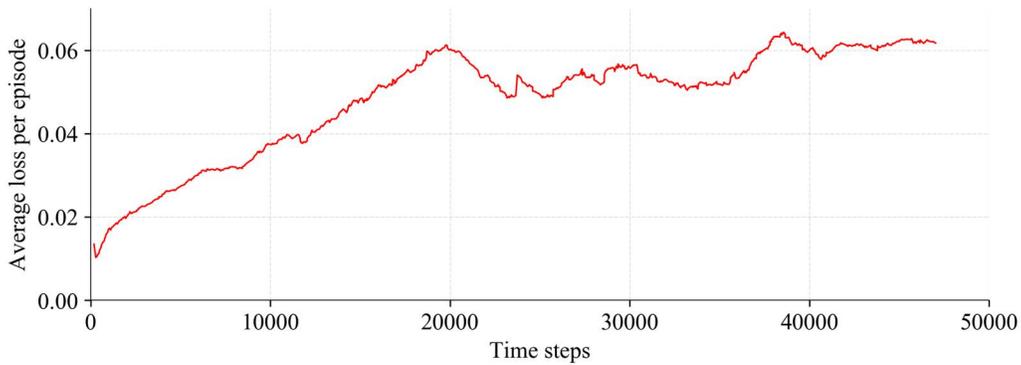

(a)

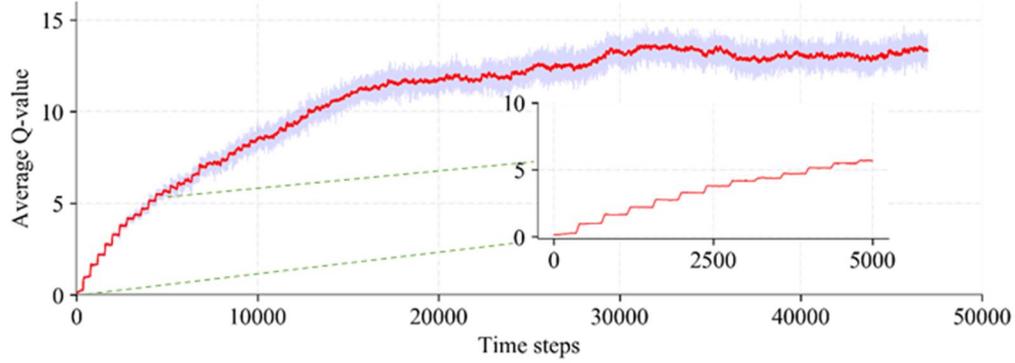

(b)

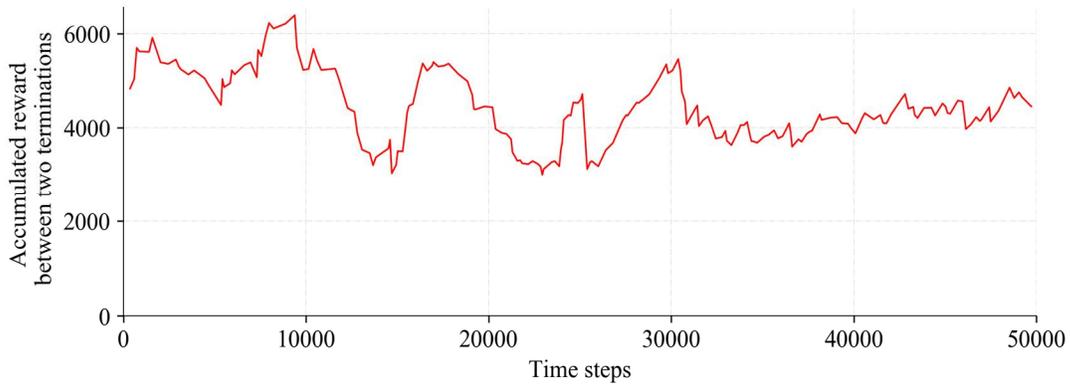

(c)

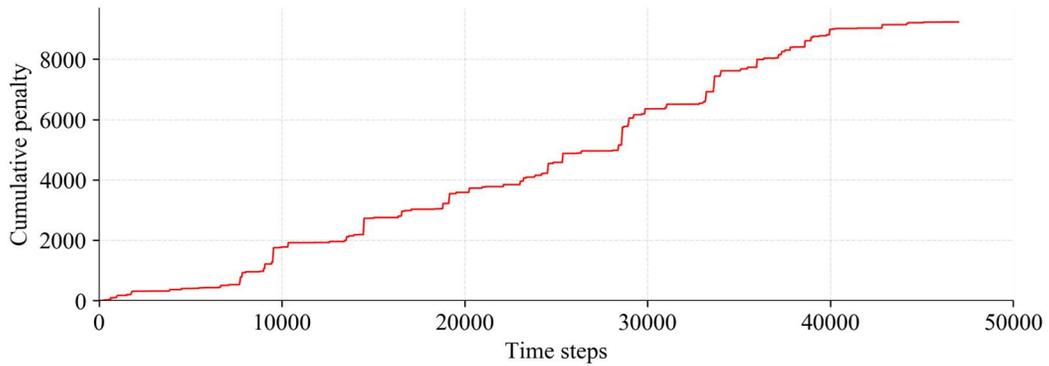

**(d)**

**Fig. 5.** Evolution of (a) average loss per episode; (b) average Q-value; (c) accumulated reward between two terminations; and (d) cumulative penalty in the course of DQN training

To further show the effectiveness of the DQN algorithm training, we apply the DQN algorithm throughout its training to three randomly generated problem instances of the same size (50 requests and 22 crowdsourcees). Fig. 6 shows the TSC results when applying the DQN algorithm with the most up-to-date DNN weight parameters every 2,500 time steps. It can be seen that TSC will be drastically reduced after the first few hundreds of time steps. For example, for problem instance 1 TSC reduces by more than three-quarters from 1,200 to less than 300. Afterwards, the improvement in TSC is more incremental with some rebounds. Consistent with Fig. 5(d), after 40,000 time steps, TSC becomes very stable across all three problem instances.

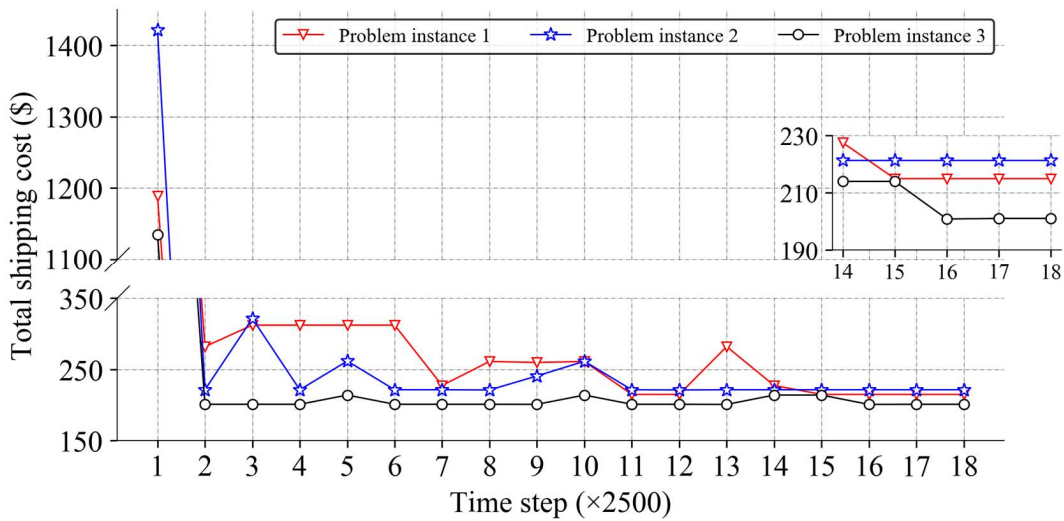

**Fig. 6.** Evolution of total shipping cost during training

### 4.1.3  Assessing the benefits of heuristics-guided action choice

Recall that one novelty of our proposed DRL algorithm lies in the embedment of heuristics-guided action choice in DRL. At each time step, the DRL agent performs one of the five types of actions to create new or change existing crowdsourcee routes. To compare the proposed DRL algorithm with a DRL algorithm without heuristics-guided action choice, a neighborhood move will be randomly chosen given any of the first four action types, as described in Appendix. Similar to what we do in Fig. 6, we apply the DQN algorithm throughout its training to two randomly generated problem instances and present the TSC results using the most up-to-date DNN weight parameters every 2,500 time steps. For each problem instance, we train the DQN algorithm twice, one with heuristics-guided action choice and the other without. The results are shown in Fig. 7.

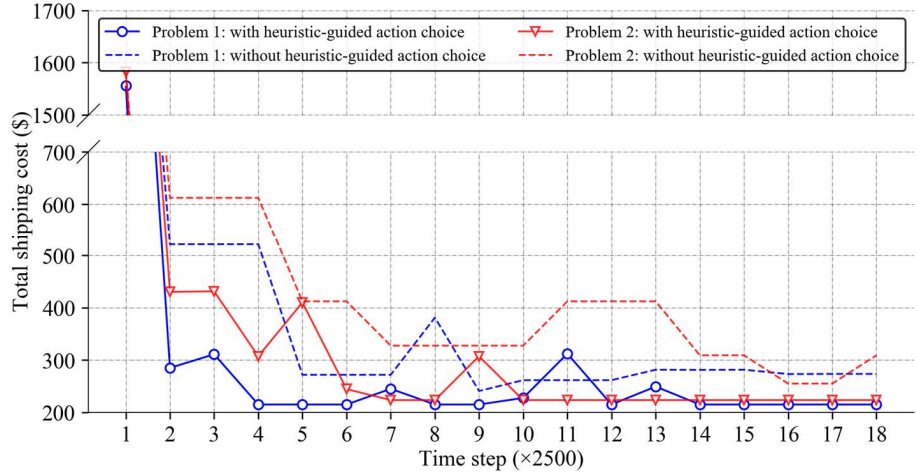

**Fig. 7.** Comparison of total shipping cost with and without heuristics-guided action choice during training

For the first problem instance (in blue), we observe that at the beginning of the training, the TSC curve without heuristics-guided action choice is slower than the TSC curve with heuristics-guided action choice. Throughout the training, the TSC curve without heuristics-guided action choice remains well above the TSC curve with heuristics-guided action choice. For the second problem instance (in red), the advantage of heuristics-guided action choice is even more visible since the convergence to the lowest stable TSC is sooner. Additionally, at the end of the training, a substantial TSC gap remains. In fact, the final TSC without heuristics-guided action choice is 38.9% higher than the final TSC with heuristics-guided action choice. Overall, the results clearly demonstrate the advantage of heuristics-guided action choice in DQN training.

Fig. 8 presents comparisons of applying the trained DRL models to 20 randomly generated problem instances. The reduction of TSC with heuristics-guided action choice is clearly observed. Across the 20 problem instances, the average TSC reduction is 15.8% with a standard deviation of 7.5%. The largest reduction, which occurs to problem instance 6, is 28.9%.

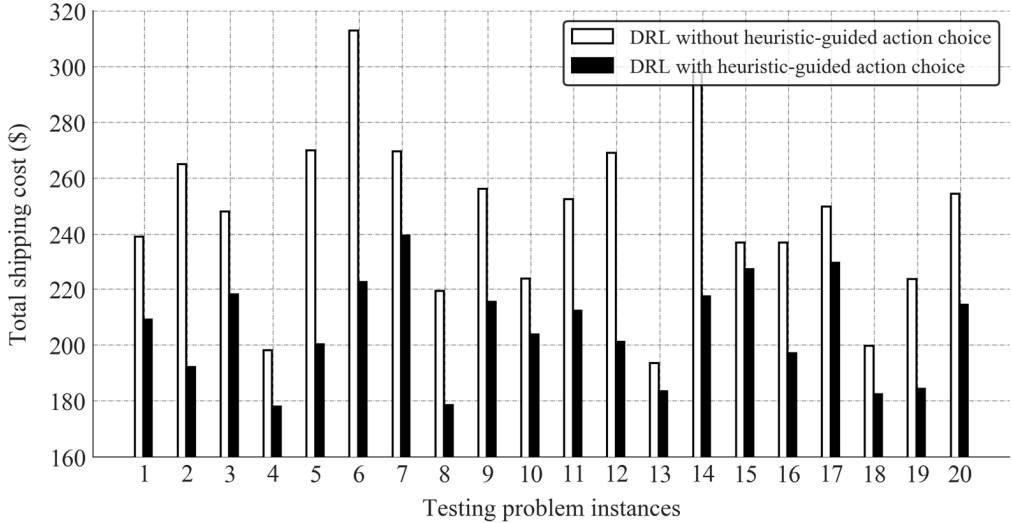

**Fig. 8.** Comparison of total shipping cost with and without heuristics-guided action choice during testing

Fig. 9 reports further the DQN training time without and with heuristics-guided action choice. To make sensible comparisons, we let training without heuristics-guided action choice run same number of time steps. The results show that the training time with heuristics-guided action choice (57.3 minutes) is much larger than without heuristics-guided action choice (23.2 minutes). This suggests that a non-trivial amount of added computation is needed during training for heuristics-guided action choice in order to achieve lower TSC.

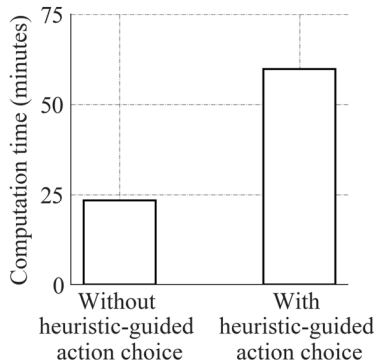

**Fig. 9.** Comparison of training time with and without heuristics-guided action choice

### 4.1.4 Assessing the benefits of rule-interposing

In this subsection we evaluate the benefits of another novelty of the proposed DRL algorithm – the integration of rule-interposing into DRL training and implementation. As in Fig. 7, we apply the DQN algorithm throughout its training to the same two randomly generated problem instances, and present the TSC results using the most up-to-date DNN weight parameters every 2,500 time steps. For each problem instance, we train the DQN algorithm twice, one with rule-interposing and the other without. The results are shown in Fig. 10.

For the first problem instance (in blue), although the TSC without the two rules appears to be diminishing at the beginning of the training, the TSC value rebounds after 15,000 time steps, then declines and meets the TSC curve when the two rules are used at around 32,500 time steps. Afterwards, the TSC curve without the two rules experiences some fluctuations and remains well above the TSC curve with the two rules. For the second problem instance (in red), the advantage of rule-interposing is even more visible since the convergence to the lowest stable TSC is much sooner. At the end of the training, a substantial TSC gap remains: the final TSC without the two rules is around 26.9% higher than the final TSC with the two rules. Overall, the results clearly demonstrate the advantage of rule-interposing in DQN training.

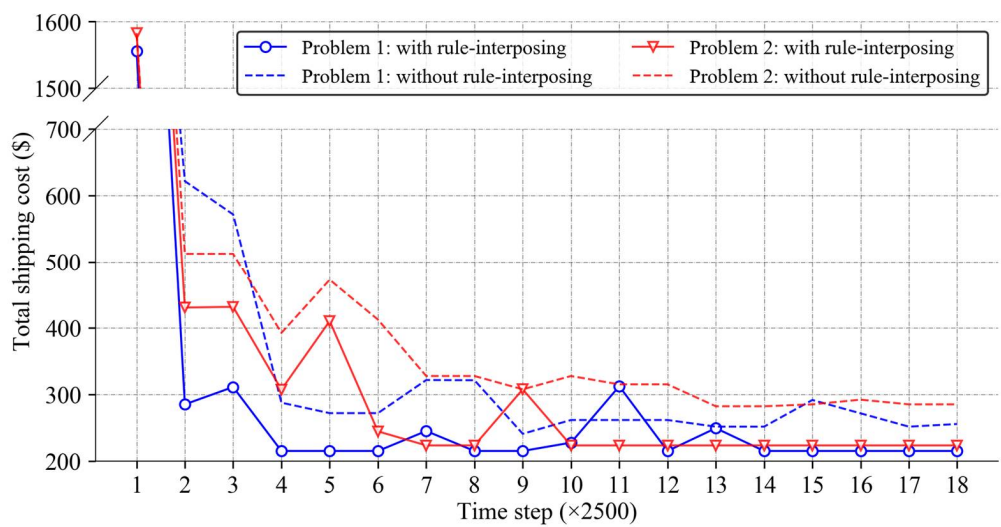

**Fig. 10.** Comparison of total shipping cost with and without rule-interposing

Fig. 11 presents comparisons of applying the trained DRL models to the same 20 randomly generated problem instances as in subsection 4.1.3. We observe an overall reduction of TSC with rule-interposing. The average TSC reduction across all 20 instances is 3.6% with a standard deviation of 5.85%. The largest reduction, which occurs to problem instance 6, is 24.1%.

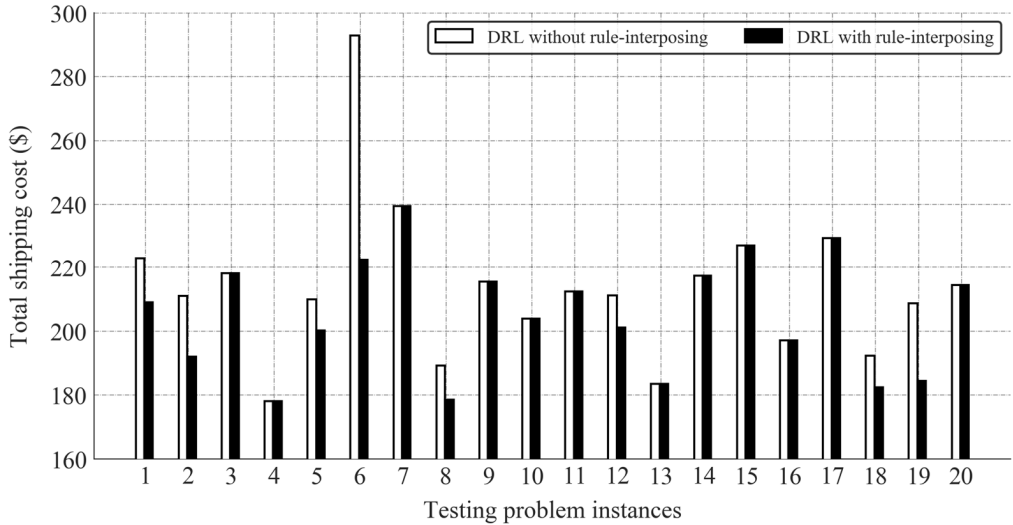

**Fig. 11.** Comparison of total shipping cost with and without rule-interposing during testing

Fig. 12 reports further the DQN training time with and without the two rules embedded. Again, to make sensible comparisons, we allow the training without the two rules to run the same number of time steps. The results show that the training time with rule-interposing (57.3 minutes) is slightly higher than without rule-interposing (52.7 minutes), which is associated with additional computation to discourage repeated visit of routes or node sequences during neighborhood moves.

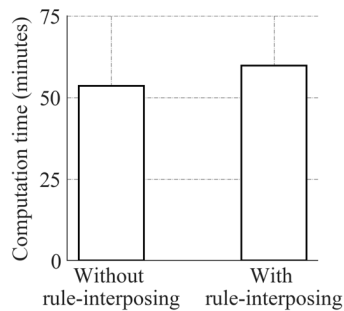

**Fig. 12.** Comparison of training time with and without rule-interposing

### 4.1.5 Comparison with heuristic methods

To further gauge the performance of the DRL-based approach, this subsection compares the proposed approach with three popular heuristics: simple heuristic, reactive Tabu search (RTS), and simulated annealing (SA). The simple heuristic basically performs Steps 1-2 of the insertion action described in subsection 3.2.3.1, and can generate solutions very fast. However, it does not explore neighborhood moves. Therefore, the resulting solution can be far from optimum. RTS is a hierarchical heuristic that dynamically adjusts search parameters and alternates between different neighborhoods while seeking the optimal routing solution, based on the state and quality of the search. Our implementation of RTS follows Nanry and Barnes (2000) with consideration of three types of neighborhood moves (intra-route move, inter-route move, and 1-exchange). SA is based on the analogy between the simulation of solids annealing and the problem of solving large combinatorial optimization problems (Kirkpatrick et al., 1983; van Laarhoven and Aarts, 1987). Prior research shows that SA can yield reasonably good solutions for large VRP instances and can be faster than other heuristics such as Tabu search and genetic algorithm (Tan et al., 2001). At each temperature during cooling, an intra-route move, an inter-route move, and an 1-exchange move as described in Ahamed and Zou (2020) are performed in sequence, with each move followed by an evaluation that accepts not only an improved solution, but also an inferior solution with certain probability. The parameter setting of SA follows those in Kafle et al. (2017).

For the simple heuristic, it terminates when all feasible insertions of requests are performed. In implementing RTS and SA, we allow for a sufficient number of iterations until the reduction in TSC is not visible (TSC change is less than 2% in the last ten iterations). Fig. 13 presents the TSC results using DRL and the three heuristics, for 20 randomly generated problem instances. DRL yields the best solution in 17 out of the 20 problem instances. In contrast, the solutions using the simple heuristic are much worse, despite small computation time as shown in Fig. 14. On the other hand, while the TSC results from RTS and SA are closer to those using DRL, the computation time is much longer, by more than an order of magnitude (10-30 minutes vs. around 1 second). Considering both solution quality and time, the comparison clearly indicates the superiority of DRL.

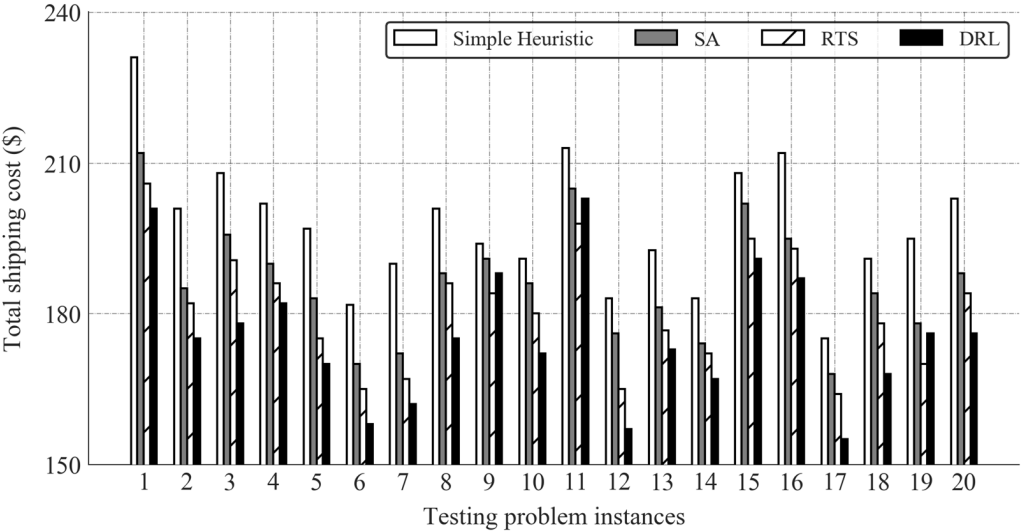

**Fig. 13.** Comparison of DRL with existing heuristics in terms of TSC (medium-size problems)

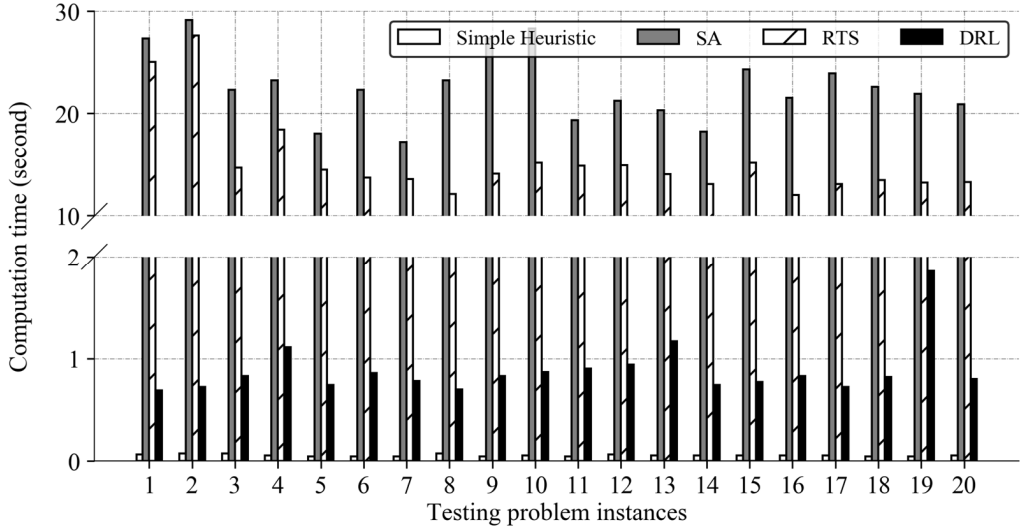

**Fig. 14.** Comparison of DRL with existing heuristics in computation time (medium-size problems)

### 4.1.6 Sensitivity of DQN training to hyperparameter values

Finally, we investigate the sensitivity of DQN training to the values of four hyperparameters: (a) decay rate $\xi$; (b) learning rate $\alpha$; (c) discount factor $\gamma$; (d) target network update frequency $\delta$. Fig. 15 presents the results. In each graph in Fig. 15, a curve corresponds to a specific value of the hyperparameter under investigation and is obtained in a similar fashion as the curves in Fig. 6, for a randomly generated problem instance. For a given graph, the three other hyperparameters not investigated take their values in Table 4. While Fig. 15 reports TSC values of one problem instance, we have also experimented with many other randomly generated problem instances and found consistent results. It can be seen that, for all graphs in Fig. 15, the chosen value for each hyperparameter produces more stable TSC curves than the alternative values. In addition, the final TSC using the chosen hyperparameter value is always smaller than using alternative values. Thus, the results reaffirm our choice of the hyperparameter values.

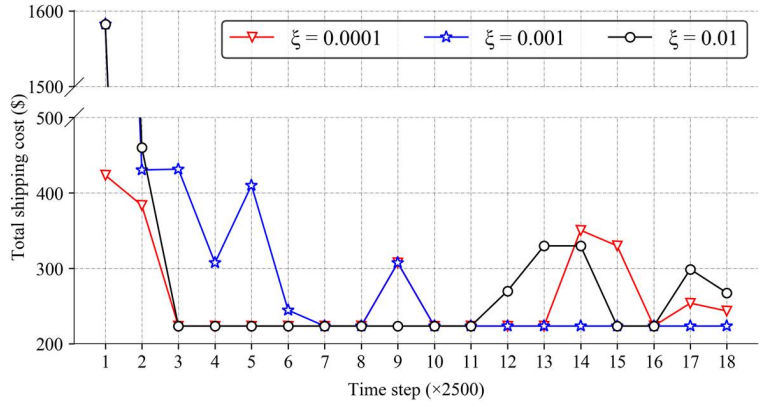

(a)

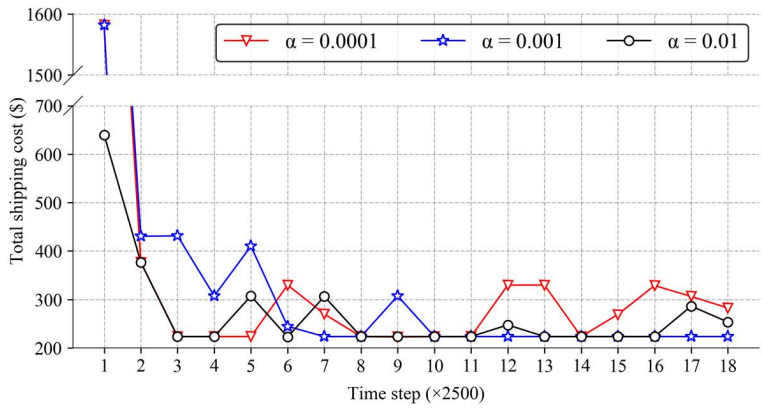

(b)

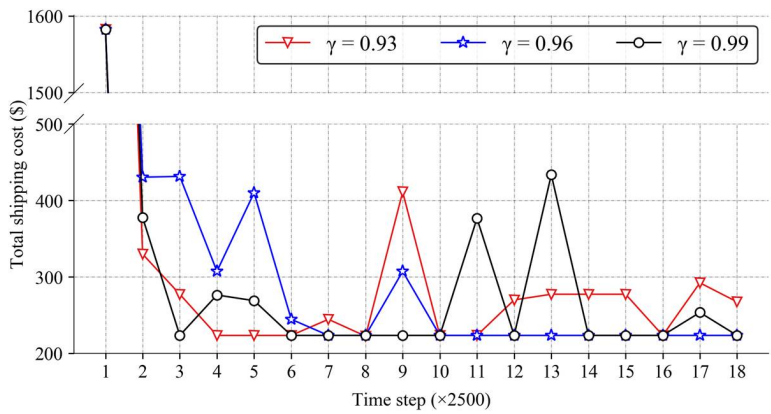

(c)

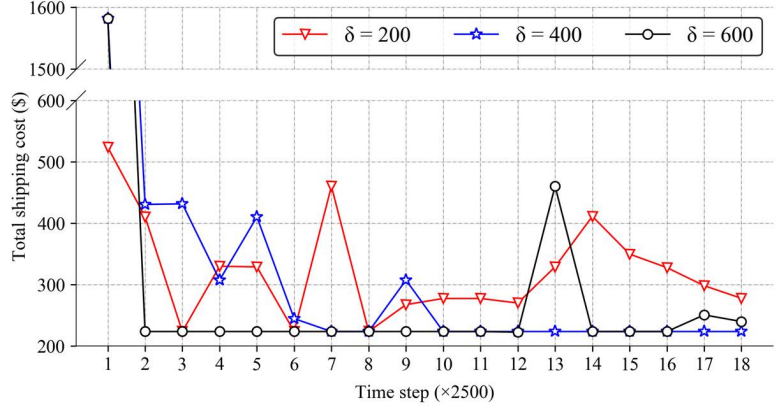

(d)

**Fig. 15.** Sensitivity of total shipping cost to different hyperparameter values: (a) $\xi$; (b) $\alpha$; (c) $\gamma$; (d) $\delta$

## 4.2 Larger-size problems

### 4.2.1 Setup

The larger-size problem instance considers problems of assigning 200 requests to 70 crowdsourcees, which are of comparable size to many pickup-and-delivery operation planning problems investigated in the existing literature (Braekers and Kovacs, 2016; Yu et al., 2019; Ghilas et al., 2016; Shao et al., 2020). Apart from a larger number of requests and crowdsourcees, other setups are the same as in the medium-size problems. With a larger problem size, it is natural to expect a higher number of time steps per episode to insert all requests and perform neighborhood moves of the requests. Therefore, we increase the length of an episode to 300 time steps. Following a similar informal search as in subsection 4.1.1, the penalty parameters in the reward specification are set to be $\vartheta = 0.25$, $\tau = 0.15$, and $\rho\phi = 0.2$, and the length of Tabu tenure to be 12 subsequent actions. The episode termination threshold $\mathcal{K}$ is decreased to -175. Decay rate $\xi$ is set as 0.002. Other hyperparameter values remain the same. The training time takes about 3 hours and 10 minutes.

### 4.2.2 Comparison of solutions using DRL and heuristics

Similar to subsection 4.1.5, this subsection compares the performance of the DRL-based approach with the three heuristics (simple heuristic, RTS, and SA). 20 problem instances with 200 requests and 70 crowdsourcees are randomly generated. Fig. 16 shows that DRL yields the best solution in 18 out of the 20 instances. As in subsection 4.1.5, the solutions from the simple heuristic are always the worst, despite small computation time (Fig. 17). While the resulting TSC values from RTS and SA are closer to those from DRL, the computation time is much longer (around 15-20 minutes, as opposed to 6-10 seconds using DRL). By comparing the computation time change from the medium-size problem (Fig.

14), it is clear that DRL is much more scalable than RTS or SA for solving the crowdshipping problems in this paper.

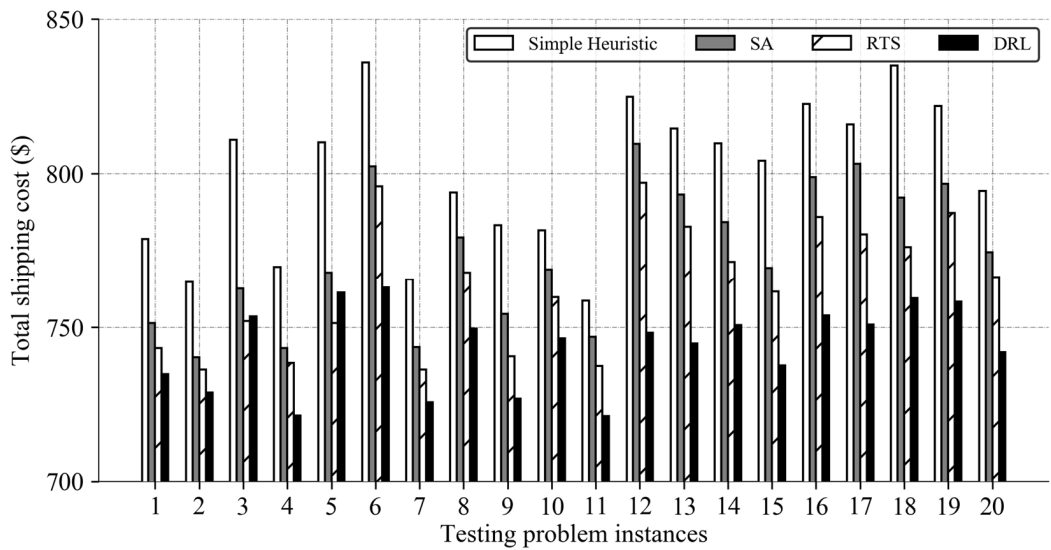

**Fig. 16.** Comparison of DRL with existing heuristics in terms of TSC (larger-size problems)

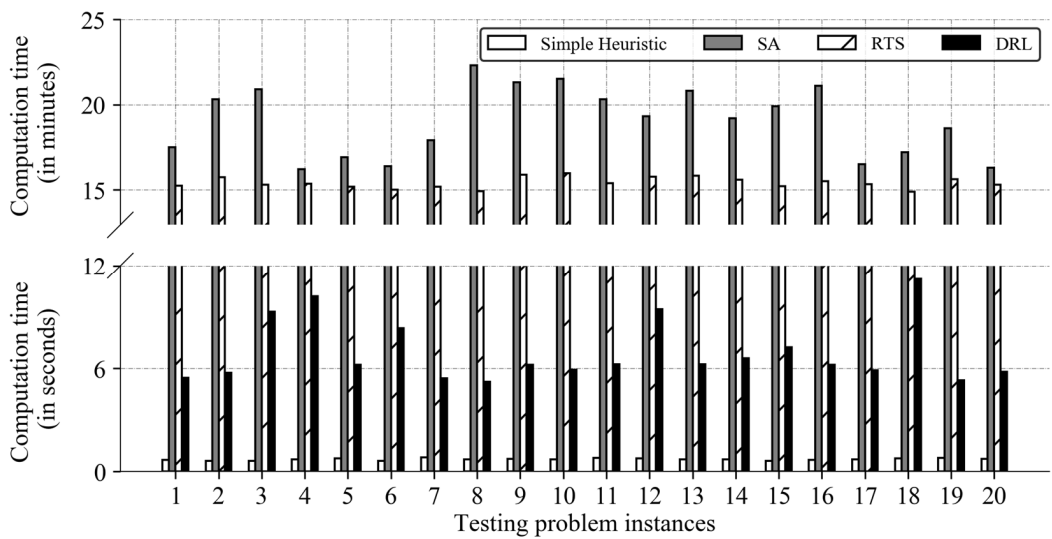

**Fig. 17.** Comparison of DRL with existing heuristics in computation time (larger-size problems)

## 5  Conclusion

Crowdshipping has gained increasing popularity for urban delivery given the low cost of hiring *ad hoc* couriers to perform pickups and deliveries. In this paper, we propose a novel, deep reinforcement learning-based approach to seek high-quality and computationally efficient assignment of requests to crowdsourcees. In performing the assignment, we consider that requests have time windows for pickup and delivery. In addition, crowdsourcees have limited time availability and carrying capacity. The novelty of the proposed DRL approach lies in its new characterization of system states, the embedment of heuristics-guided action choice, and the integration of rule-interposing into DRL training and implementation. The effectiveness of the approach is demonstrated through extensive numerical analysis. The results show the benefits brought by the heuristics-guided action choice and rule-interposing in DRL training, and the superiority of the proposed approach over existing heuristics in both solution quality, time, and scalability.

With its comprehensive and detailed specifications of states, actions, and rewards, the proposed approach not only has the potential to improve the efficiency of crowdshipping operation planning, but provides a new avenue and generic framework that could be adopted to other pickup and delivery problems and vehicle routing contexts. For example, another type of crowdshipping with all requests originating from a central location (depot) can be viewed as a special case of the problems investigated in this paper. Also, while we consider dedicated crowdsourcees in the paper, the proposed DRL-based approach can be conveniently adapted to the context of opportunistic crowdsourcees given the origin and destination of the original trip of each crowdsourcee.

For possible extension of the proposed approach, a few directions are suggested. First, future efforts could be made to investigate a dynamic version of the problem. In this case, the choice of actions should not only consider the present situation but also look ahead. Second, in the real world the pickup and delivery locations of shipping requests are usually in different spatial distributions (e.g., the locations of restaurants/retail stores in a city may be quite different from the locations of residential buildings), which gives rise to the need for proactively relocating idle crowdsourcees to balance the spatial distribution of crowdsourcee supply and request pickup demand. It will be interesting to explore how to incorporate relocation decisions in the DRL framework. Third, some behavioral aspects, e.g., a crowdsourcee rejects an assigned request, could be added to further enrich the flexibility of the DRL model.

# Acknowledgment

This research is funded by the National Science Foundation under Grant Number 1663411. The financial support of the National Science Foundation is gratefully acknowledged. An earlier version of the paper was presented at the INFORMS 2020 Annual Meeting. We also thank the attendees of the presentation for providing their helpful feedback.

# Appendix: Identification of the specific action to take given the action type under a DRL algorithm without heuristics-guided action choice

| **Insertion** |  |
|---|---|
| **Step 1:** | *Select a request.* |
|  | Among the unassigned requests, randomly select an unassigned request. |
| **Step 2:** | *Insert the request to a route.* |
|  | Insert the request to the end of a randomly picked crowdsourcee route (which can be an existing or a new route). If the insertion is not feasible, then randomly pick another crowdsourcee route. If a feasible insertion cannot be found, then do nothing. |
| **Intra-route move** |  |
| **Step 1:** | *Select a route.* |
|  | Select the crowdsourcee route with the largest remaining available time (based on Rule 1 in subsection 3.4.1). |
| **Step 2:** | *Move a request from the route to a different location on the same route.* |
|  | Randomly pick a request from the route. Enumerate all feasible moves of the pickup and delivery nodes of the request on the route. Pick the move with the maximum reduced cost. If such a move does not exist, then randomly pick another request and do the same. If such a move cannot be found after enumerating all requests on the route, then do nothing. |
| **Inter-route move** |  |
| **Step 1:** | *Select a request.* |
|  | Select the crowdsourcee route with the largest occupation time (based on Rule 1 in subsection 3.4.1). |
| **Step 2:** | *Move the request to the end of a different route.* |
|  | Randomly select a request from the route. Investigate moving the request to the end of a different route that is also randomly picked. If the move if feasible, perform the move. Otherwise, randomly pick another route and investigate moving the request to the end of the route. If the request cannot be moved to the end of any different route, then do nothing. |
| **1-exchange** |  |
| **Step 1:** | *Select two routes.* |
|  | Select the two crowdsourcee routes with the largest and the second largest unused service time (based on Rule 1 in subsection 3.4.1). |
| **Step 2:** | *Select requests from the two routes and exchange.* |
|  | Randomly select a request from each routes and exchange their locations. |

Note that for intra-route move, inter-route move, and 1-exchange, we do not consider Rule 2 of subsection 3.4.2 since the rule is related to heuristics-guided action choice.